\documentclass[journal]{IEEEtran}
\usepackage{amsthm,amsfonts,amssymb}
\usepackage{graphicx}
\usepackage{bbding}
\usepackage{color}
\usepackage{subfigure}
\newcommand{\new}[1]{\textcolor{black}{#1}}

\usepackage{multirow}
\usepackage{booktabs}
\usepackage{fancyhdr}
\usepackage{amsmath}
\usepackage{amsfonts}
\usepackage{graphicx}
\usepackage{rotating}
\usepackage{pifont}
\begin{document}
\newcommand{\gr}[1]{\textcolor{green}{#1}}
\title{Language-Assisted Human Part Motion Learning for Skeleton-Based Temporal Action Segmentation}

\author{Bowen~Chen,
	Haoyu~Ji,
 Zhiyong Wang,
	Benjamin Filtjens,
 Chunzhuo~Wang,
	Weihong~Ren,
	\\
	Bart Vanrumste,~\IEEEmembership{Senior Member,~IEEE}
	and~Honghai~Liu$^*$,~\IEEEmembership{Fellow,~IEEE}% 
	\thanks{
* corresponding author

Bowen Chen, Haoyu Ji, Zhiyong Wang, Weihong Ren and Honghai Liu are with the State Key Laboratory of Robotics and Systems, Harbin Institute of Technology Shenzhen, Shenzhen 518055, China (e-mail: chenbw@stu.hit.edu.cn,  honghai.liu@icloud.com). Benjamin Filthens, Chunzhuo Wang and Bart Vanrumste are with the e-Media Research Lab, Department of Electrical Engineering (ESAT), KU Leuven, Leuven, Belgium. 
Bowen Chen is also with the e-Media Research Lab, KU Leuven, Leuven, Belgium.
Honghai Liu is also with the Peng Cheng Laboratory, Shenzhen 518000.
        }
 }
\markboth{Journal of \LaTeX\ Class Files,~Vol.~14, No.~8, August~2015}%
{Shell \MakeLowercase{\textit{et al.}}: Bare Demo of IEEEtran.cls for IEEE Journals}

\maketitle
	
\begin{abstract}
	Skeleton-based Temporal Action Segmentation involves the dense action classification of variable-length skeleton sequences. Current approaches primarily apply graph-based networks to extract framewise, whole-body-level motion representations, and use one-hot encoded labels for model optimization. However, whole-body motion representations do not capture fine-grained part-level motion representations and the one-hot encoded labels neglect the intrinsic semantic relationships within the language-based action definitions. 
    To address these limitations, we propose a novel method named Language-assisted Human Part Motion Representation Learning (LPL), which contains a Disentangled Part Motion Encoder (DPE) to extract dual-level (i.e., part and whole-body) motion representations and a Language-assisted Distribution Alignment (LDA) strategy for optimizing spatial relations within representations.
    Specifically, after part-aware skeleton encoding via DPE, LDA generates dual-level action descriptions to construct a textual embedding space with the help of a large-scale language model. Then, LDA motivates the alignment of the embedding space between text descriptions and motions. This alignment allows LDA not only to enhance intra-class compactness but also to transfer the language-encoded semantic correlations among actions to skeleton-based motion learning. Moreover, we propose a simple yet efficient Semantic Offset Adapter to smooth the cross-domain misalignment. Our experiments indicate that LPL achieves state-of-the-art performance across various datasets (e.g., +4.4\% Accuracy, +5.6\% F1 on the PKU-MMD dataset). Moreover, LDA is compatible with existing methods and improves their performance (e.g., +4.8\% Accuracy, +4.3\% F1 on the LARa dataset) without additional inference costs. 
	
\end{abstract}

% Note that keywords are not normally used for peerreview papers.

\IEEEpeerreviewmaketitle

\section{Introduction}
% SAS background
\IEEEPARstart{T}{emporal} Action Segmentation (TAS) aims to partition a continuous, long sequence of actions into multiple action segments~\cite{ASReview}. Unlike action recognition, TAS aims to accurately localize action transitions in varying length sequences, making it ideal for fine-grained analyses in complex, real-world scenarios. Depending on the input modality, TAS methods can be divided into two categories: video-based action segmentation (VTAS) and skeleton-based action segmentation (STAS). VTAS takes RGB frames as input, \new{which usually rely} on offline \new{dense} feature pre-extraction and are easily affected by the quality of the features.  In contrast, STAS methods utilize skeleton-related signals like IMU and joint coordinates, which are low-dimensional and robust against background noise and lighting variations~\cite{10464339}.
\begin{figure}
	\centering
	\includegraphics[width=0.5\textwidth]{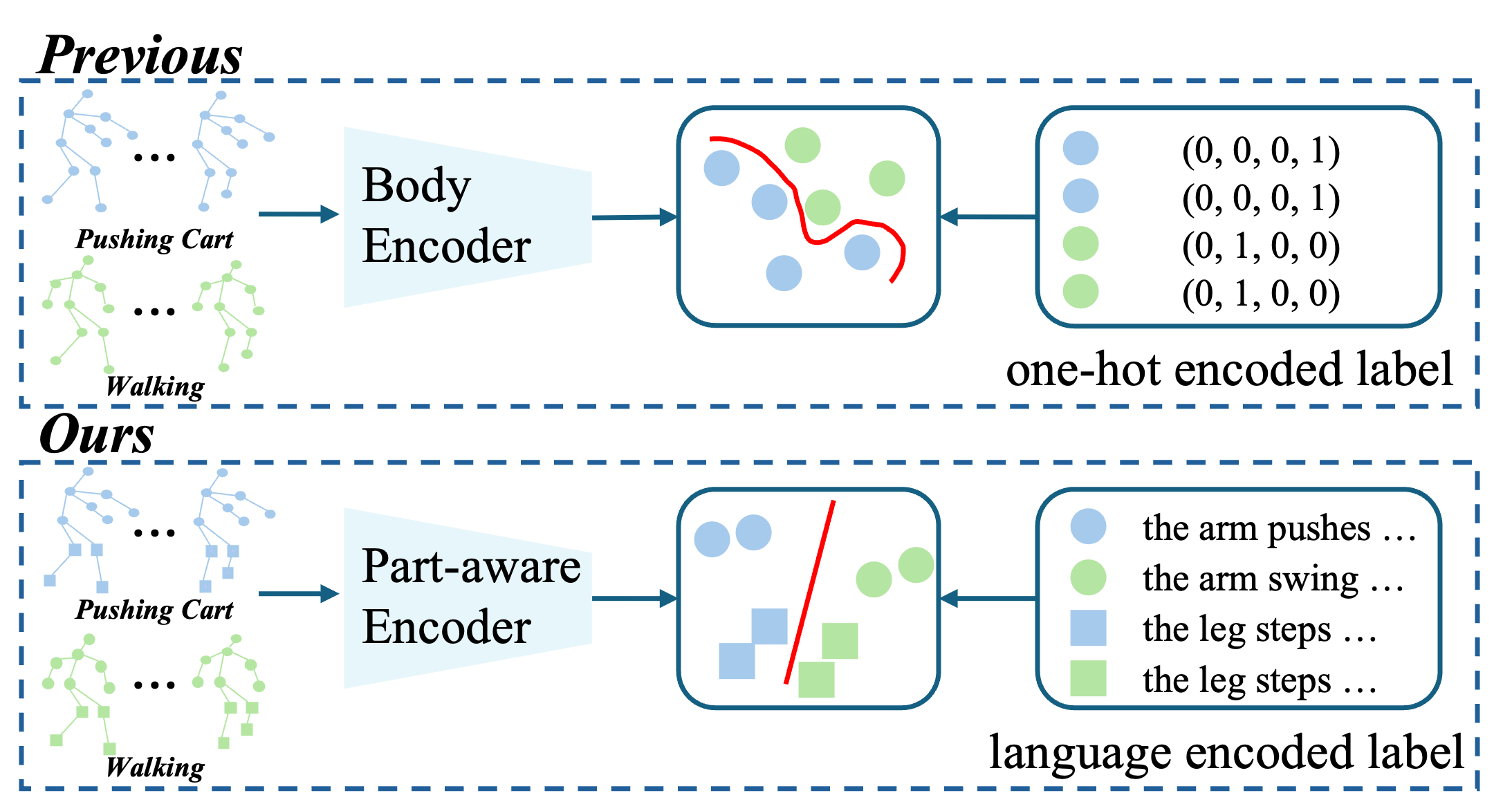}
	\caption{ \label{fig:fig0}Comparison between previous STAS methods and ours. Previous methods extracts global (whole-body) features and optimized the motion representations only with one-hot encoded label, losing the part-level motion understanding and semantic correlations across actions. Our method enhances action understanding at the part-level and \new{integrates semantic
correlations as auxiliary knowledge} into motion representation learning, achieving a deeper comprehension aligned with human cognition.}
\end{figure}

Accurate action segmentation hinges on learning high-quality spatio-temporal motion representations. 
Early STAS methods~\cite{okeyo2014dynamic} relied on hand-crafted feature extraction, which had limited performances. With the development of deep neural networks and the emergence of large-scale skeleton-based motion datasets~\cite{shahroudy2016ntu}, current works typically use graph convolutional networks (GCN)~\cite{msgcn, STGA-Net} to extract motion representations. \new{These networks often employ a coupled spatio-temporal pipeline, where multiple spatial layers aggregate features across joints, and temporal layers capture movement trajectories across frames. After that, segmentation are achieved through classification of the dense representations. The whole network is trained with one-hot labels, where only class identification information is preserved~\cite{pang2019rethinking}. Most recently, several studies noted that this coupled pipeline can lead to over-smoothed features~\cite{li2023involving,DeST}. To address this issue, a novel decoupled framework, named DeST~\cite{DeST}, is proposed and achieves significant performance improvements. }

Despite of achieved success, we argue that current methods neglects part-level motion understanding and overlooks the rich semantic correlations that are embedded in language-encoded action definitions, as illustrated in Fig.~\ref{fig:fig0}.
For instance, in the action \textit{Pushing Cart}, while the \textit{arm} (a joints group that consists of the hand, wrist, and elbow) exhibit consistent \textit{pushing}, the \textit{leg} usually performs \textit{stepping forward and backward}. This distinct motion consistency within and independence across body parts are crucial yet \new{is insufficiently emphasized in existing STAS methods. Although several studies have explored part-level modeling in other action understanding tasks~\cite{dang2021msr,rao2021sm}, their spatio-temporal modeling pipeline fails to meet the discriminative feature requirements necessary for fine-grained segmentation~\cite{DeST}. Achieving part awareness while keeping motion feature discriminative deserve further exploration.} Nevertheless, the action \textit{Pushing Cart} and \textit{Walking} should share similar motion patterns of the \textit{leg}, while significant differences exist of the \textit{arm}. This semantic correlation can be described clearly with human language: ``both actions involves the leg stepping forwards and backwards, but the arm motions differ". In other words, the label \textit{Pushing Cart} contains highly abstract latent knowledge encoded by human language, which is neglected by one-hot encoding. In this case, the model must learn a complex decision boundary in representation space~\cite{pang2019rethinking}, leading to a poor generalization and insufficient understanding of actions. 

\new{To address these limitations, this paper introduces a new STAS method, called \textbf{L}anguage-assisted human \textbf{P}art motion representation \textbf{L}earning (LPL). LPL consists of a Disentangled Part motion Encoder (DPE) for part-aware motion encoding and a Language-assisted Distribution Alignment (LDA) for motion representation space refinement. DPE decouples human modeling into two phases, independent decoupled spatio-temporal interaction for each body part, followed by cross-part information exchange through an inter-part interaction module. This design leverages the advantages of DeST to avoid excessive feature smoothing while achieving part awareness. After DPE encoding, LDA reshapes the motion feature space structure through text-motion feature space distribution alignment, transferring text-encoded prior knowledge to the motion domain via part-level neighbor relationships. With LDA, the motion representation achieves 1) intra-class compactness and inter-class separability, leading to clearer decision boundaries, and 2) the neighbor relationships among clusters are determined by the high-level semantics of labels, aligning more closely with human definitions of actions.}
Additionally, we also include an efficient adaption module to
learn the semantic offset between skeleton and text and thereby
reduce the cross-domain misalignment.

Experiments conducted on three public datasets covering various scenarios, such as traffic control and daily activities, validate the superior performance of our approach. The proposed LDA successfully built a well-constructed motion representation space and owns compatibility on improving current STAS methods without additional \new{computations}. To our knowledge, it is the first work to introduce part-level semantic understanding into STAS methods.
Our contributions are three-fold:
\begin{itemize}
	\item We propose a novel disentangled graph convolutional network for part-aware motion representation extraction, which avoids spatio-temporal feature over-smoothing;
	\item A language-assisted distribution alignment strategy is incorporated, which efficiently transfers the language-encoded action distribution to refine the structure of the motion representation space;
	\item We propose a simple yet efficient semantic offset adapter to mitigate the cross-domain misalignment;

\end{itemize}

The remainder of this paper is organized as follows. Section II reviews previous works on action understanding and applications of language models. In Section III, we introduce our method, and in Section IV conduct experiments on public datasets. Finally, we conclude the paper in Section V.

\section{Related Work}
The challenges in Temporal Action Segmentation (TAS) predominantly involve two aspects: modeling the spatio-temporal motion representations at the frame level, and understanding the action context. This section reviews recent advancements in TAS methods focusing on these challenges, along with the related applications of graph-based skeleton modeling and large-scale language models in action understanding tasks.
\subsection{ Video-based Temporal Action Segmentation}
Video-based Temporal Action Segmentation (VTAS) processes RGB video frames as inputs. Given the high redundancy inherent in images, optimizing a large volume of video frames for neural network inputs presents significant challenges. Therefore, VTAS often employs large pre-trained action recognition models, such as I3D~\cite{i3d}, to serve as offline feature extractors for framewise feature extraction. Subsequent modules engage in temporal modeling to derive dense motion representations for classification. Initial classifications typically exhibit considerable noise, particularly manifesting as over-segmentation errors, necessitating temporal refinement stages to smooth the action predictions by integrating action context\new{~\cite{farha2019ms}}. Recent research has primarily concentrated on developing modules for temporal modeling and refining strategies.

Dilated Temporal Convolutional Networks (Dilated-TCN) have emerged as the predominant tool for temporal modeling due to their flexibility in adapting to various input lengths by adjusting the dilation rate~\cite{lea2017temporal,lei2018temporal}. Besides, graph-based methods \cite{ghosh2020stacked, huang2020improving} and the Transformer architecture~\cite{asformer} also show excellent performance. In terms of prediction refinement, current methods typically use a multi-stage optimization strategy to suppress noise, taking the probability distribution of initial predictions as input for reasoning about the potential sequence logic of actions~\cite{MS-TCN}. Additionally, studies from~\cite{wang2020boundary, ishikawa2021alleviating} explicitly incorporate action boundary predictions to refine the action predictions. These methods significantly reduce over-segmentation errors; however, undetected action boundaries can also lead to the loss of action segments. UVAST~\cite{uvast} introduces an innovative approach by employing a top-down Transformer architecture that initially predicts action segments and subsequently assigns frames to each segment. Other studies have explored network architecture search~\cite{gao2021global2local}, diffusion models~\cite{liu2023diffusion}, and multiple data sources fusion~\cite{van2023aspnet} to further improve the performance.
\subsection{Skeleton-based Temporal Action Segmentation}
\new{Current STAS methods often follow Multi-Stage Spatial-Temporal Graph Convolutional Neural Network (MS-GCN)~\cite{msgcn}}, a pioneering approach that adapts the skeleton-based action recognition backbone~\cite{yan2018spatial} to extract dense framewise representations, followed by temporal refinement stages to optimize initial predictions. Subsequent advancements have focused on enhancing the quality of framewise motion representations through techniques such as spatio-temporal attention~\cite{STGA-Net} and data augmentation~\cite{CTC}. Despite these improvements, the task gap between action segmentation and recognition remains challenging. Recent research indicate that the action segmentation requires a finer semantic granularity compared to action recognition\new{~\cite{li2023involving}}. However, existing GCNs, essentially low-pass filters, tend to produce over-smooth features after stacking multiple layers, which obscures discriminative signals crucial for fine-grained segmentation. Moreover, the cascaded spatial-temporal layers also leads to over-smooth spatio-temporal features. To address these issues, high-frequency signal capture~\cite{IDT-GCN} and decoupled spatio-temporal interaction~\cite{DeST} are proposed and have shown significant performance improvements. However, these methods focused solely on whole-body motion representation extraction, overlooking the independence and synergy of human body part movements. Moreover, the label text, representing human high-level abstract knowledge, are simplified to one-hot encoded labels during the training process. The models are only optimized based on the relationships among one-hot logits, ignoring the semantic relationships inherent in the labels~\cite{pang2019rethinking}. This results in insufficient understanding of actions and leads to unclear inter-class decision boundaries with low generalization performance. Therefore, how to utilize the semantic relationships among body parts to achieve comprehensive understanding of motions is worth exploring.

\subsection{GCN-based Skeleton-based Action Recognition}
Contemporary methods in skeleton-based action segmentation typically convert skeleton data, composed of joints and bones, into spatio-temporal graphs to model motion patterns effectively. The Spatio-Temporal Graph Convolutional Network (ST-GCN)~\cite{ST-GCN} stands as a foundational approach. It leverages graph convolutional layers to aggregate spatial features of joints while employing temporal convolutions to capture their dynamic movements over time.
Recent advances have focused on enhancing GCNs to derive more expressive spatio-temporal motion representations~\cite{tcsvt_gcn, tcsvt_gcn2}. Spatial improvements include the implementation of attention mechanisms to emphasize key joints and multi-scale joint aggregation for analyzing actions at various semantic levels~\cite{tcsvt_part}. Efforts have also been made to develop learnable topology~\cite{huang2023graph} that facilitate direct feature aggregation across joints. Regarding temporal modeling, a range of multi-scale and attention-based methods have been introduced to better accommodate actions that occur at different speeds~\cite{chen2021multi}.
Additionally, some efforts have been directed towards improving the representational structure of motion by introducing methodologies like contrastive learning losses~\cite{shu2022multi}, aimed at a well-constructed feature space that facilitates clear decision boundaries and enhances model performance in distinguishing ambiguous classes. Despite the effectiveness of these algorithms in extracting motion patterns, the features derived from these methods still exhibit over-smoothness due to the cascaded spatio-temporal interaction~\cite{DeST}. Therefore, designing a GCN specifically for action segmentation tasks remains \new{challenging}.

\subsection{Language Model Application in Action Understanding}
With the emergence of web-scale text datasets and the development of transformer-based large-scale networks, several large-scale language models such as BERT~\cite{BERT} and T5~\cite{T5} have demonstrated powerful text understanding and expression capabilities, excelling in various vision tasks with strong zero-shot learning abilities~\cite{tcsvt_language}. In action understanding, video-based action recognition methods like ActionClip~\cite{ActionCLIP} leverage the comprehension capabilities of large-scale image-text pre-trained models to frame modeling, followed by designing temporal extraction modules for spatio-temporal action understanding. Skeleton-based action recognition methods typically focus on exploring the semantic associations embedded within labels. For instance, GAP~\cite{LST} utilizes GPT-3~\cite{gpt3} to generate detailed action description texts to facilitate semantic learning. LA-GCN~\cite{LA-GCN} constructs dynamic graphs using language texts to aid in learning semantic correlations among actions and joints. In VTAS, Bridge-Prompt~\cite{Bridge-prompt} designs prompt texts from multiple perspectives and aligns them with motion representations. UnLoc~\cite{Unloc} integrates video and category text information for comprehensive modeling. These applications inspired us to explore how to leverage language model to improve the motion representation learning in STAS.

\section{Method}
\begin{figure*}
	\centering
	\includegraphics[width=0.99\textwidth]{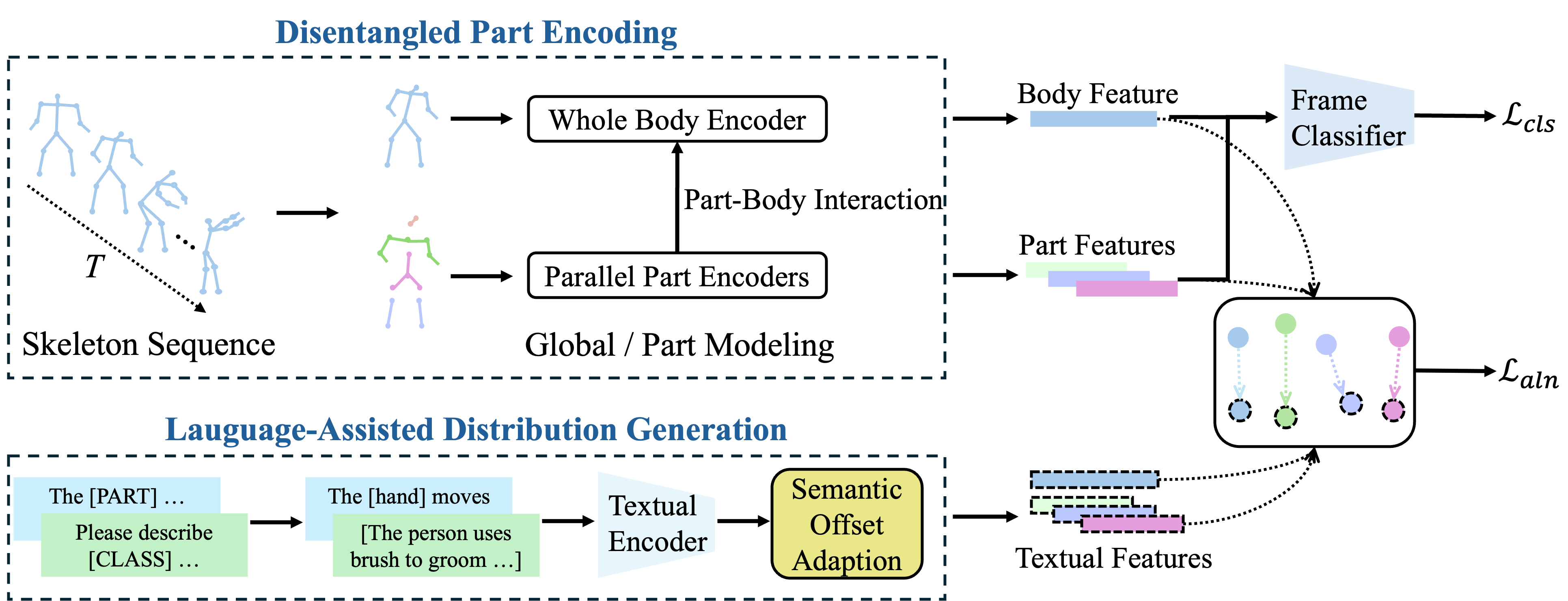}
	\caption{Framework of the proposed Language-Assisted Human Part Feature Learning (LPL). The skeleton sequence are encoded by disentangled part feature learning module (Sec. III.A). In addition to general framewise classification,  we incorporate additional language-assisted distribution learning  to generate textual feature distribution from the view of human knowledge. The contrastive loss $\mathcal{L}_{aln}$ motivates skeleton features distribution close to textual feature distribution to achieve human knowledge injection.}
	\label{fig:framework}
\end{figure*}

In this section, we present the details of the proposed LPL, as illustrated in Fig.~\ref{fig:framework}. The framework consists of two main components: 1) the disentangled encoder DPE to model part-level motion features, and 2) the language-assisted distribution alignment strategy LDA for motion representations refinement. The model is optimized by the integration of classical \new{classification loss} $\mathcal{L}_{cls}$ and \new{our proposed feature space alignment loss} $\mathcal{L}_{aln}$. In the remainder of this section, the DPE is introduced at first,  followed by the details of LDA.
\subsection{Preliminaries}
\subsubsection{General STAS Pipeline} 
Give a skeleton sequence $M\in \mathbb{R}^{C\times V\times T}$, STAS aims to identify each frame in $K$ classes, where $C$, $V$, and $T$ denote the number of channels, the number of joints, and the number of frames respectively. The $C$ depends on the modality settings, for example $C=3$ for 3D coordinates. The $V$ depends on the capture setting, for example $V=19$ for 19 IMUs in TCG~\cite{wiederer2020traffic}. \new{The joints can be grouped into $I\in N^*$ parts.} Current STAS methods utilize graph-based convolutional networks to extract motion representations, and utilize a framewise classifier to map representations into label space to generate initialized predictions. Because these predictions are usually noisy, a temporal reasoning network is commonly used for refinement.

\subsubsection{Pooling-based Part Modeling}
\begin{figure*}
	\centering
		\includegraphics[width=0.95\textwidth]{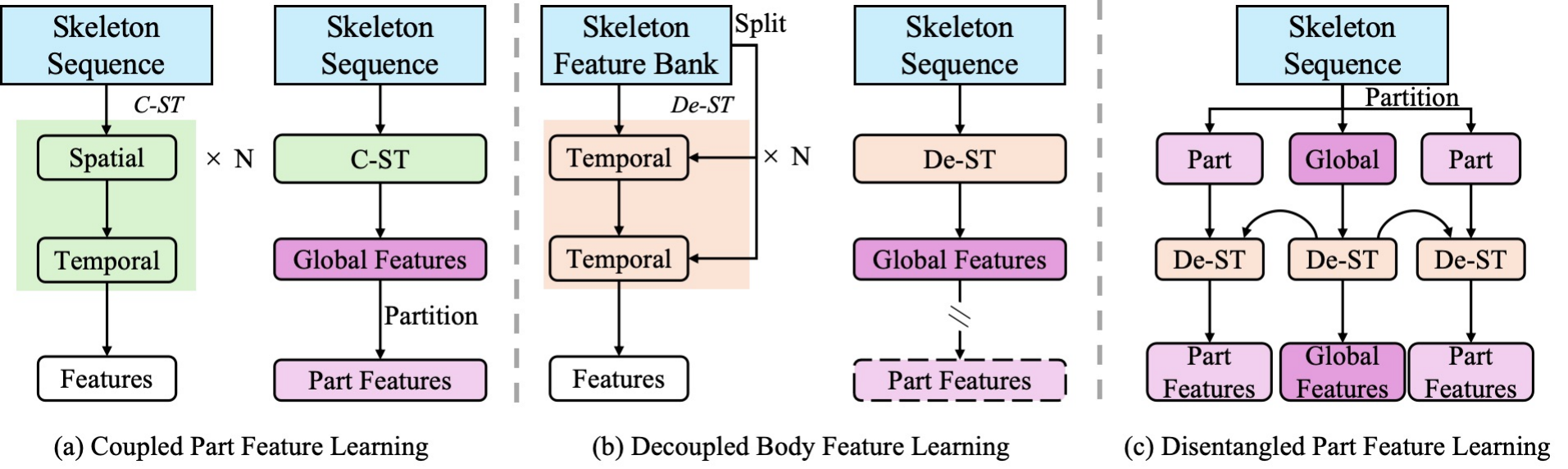}
	\caption{\label{fig:gcn}The structure comparison among our DPE and previous motion encoders. a) Traditional C-ST modeling and corresponding encoder. The part features are achieved by joints pooling over global features; b) State-of-the-art DeST modeling method, where the part modeling is absent. c) Our DPE, which extract part and body features in parallel, and keep interactions across part and whole-body. This design combines the advantages of part-awareness from C-ST and discriminative features from DeST.}
\end{figure*}

Applying part feature learning into STAS meets the challenge of feature over-smoothing. Most of STAS methods adopt topology-persist spatio-temporal modeling modules, i.e., Coupled Spatio-Temporal Framework (Coupled-ST), as illustrated in Fig.~\ref{fig:gcn}(a). They alternatively stack spatial and temporal layers to extract joint-level framewise features, and then generate part features by joints pooling~\cite{tcsvt_part}, which is widely applied in skeleton-based action recognition.  \new{Given a joint-part mapping $Q$ and body feature $f$, the $i^{th}$ part feature $f_i$ can be aggregated by:
\begin{equation}
    f_i = \text{AveragePooling}(f, Q_i),
    \label{agg}
\end{equation}
where the $Q_i$ denotes the joint indices belong to the $i^{th}$ part.}
However, the cascaded spatio-temporal interaction over-smooth motion information and producing weak spatio-temporal features~\cite{DeST}. Therefore, the Decoupled Spatio-Temporal Framework (DeST) is proposed to eliminate over-smoothing issue~\cite{DeST}, as illustrated in Fig.~\ref{fig:gcn}(b). It builds spatial feature bank through one-pass graph convolution forward, and then split spatial features into several subgroups. The model focuses on the temporal modeling while maintaining interaction with spatial subgroups. This pipeline avoids over-smoothness brought by traditional coupling modeling and significantly reduces the computation overhead because of less graph convolution operations. However, this approach abandons the human topology, leading to the absence of part feature extraction. 

\subsection{Disentangled Part Feature Learning}
To address this problem, we proposed a novel Disentangled Part Encoder (DPE), as illustrated in Fig.~\ref{fig:gcn}(c). The DPE partitions joints into parts after the spatial feature bank and extract part-level and global (whole-body) motions in parallel. At the same time, the parts and the whole-body keeps interaction layer-by-layer. This disentangled design not only benefits from discriminative features from DeST, but also keeps independence across human parts for part-level motion understanding. Next we introduce the details of DPE.

\subsubsection{Part Feature Banks} Given skeleton sequence $M$, we extract the skeleton feature bank $\mathcal{B}$ by one-pass multi-scale graph convolution. Following MS-G3D~\cite{MS-G3D}, a static multi-hop graph $A$ is defined as:
\[
	A^{(z)}_{ij} =
	\begin{cases}
		1, & \text{if } d(\alpha_i, \alpha_j) = z, z \in Z\\
		1, & \text{if } i = j,                     \\
		0, & \text{otherwise},
	\end{cases}
\]
where \(d(\alpha_i, \alpha_j)\) denotes the shortest distance (i.e., the number of edges) between joints \(\alpha_i\) and \(\alpha_j\), and the $Z$ denotes the maximum hops (13 as default).  Here the $A$ contains the semantic edges between arbitrary joints, avoiding stacking multiple graph convolution layers to extract correlation between indirectly connected joints. As a complementary, another trainable graph $\hat{A}$ is added with $A$ to capture dynamic semantic correlations among joints, performing a role of dynamic adjacent graph, which has been proven efficient in previous studies~\cite{zhang2023skeleton}. Then, the skeleton features are extracted by graph convolution operation, formulated as:
\new{\begin{equation}
	\hat{\mathcal{B}} = \text{MLP}(W^s M (\text{Concat}(\hat{A}^{(z)} + A^{(z)} ))), z\in [1,\dots,Z],
\end{equation}}
where the $\text{Concat}$ denotes the concatenation along the hop dimension, and the $\text{MLP}$ is a multi-layer perceptron for dimension adjustment.
\new{With $\hat{\mathcal{B}}$, we can aggregate part features to generate part feature banks 
$\mathcal{B}=
\left\{
\mathcal{B}_i | \mathcal{B}_i \in \mathbb{R}^{T\times V\times \vert Q_i\vert}
\right\}$ with Eq.~\ref{agg}, where $\vert Q_i\vert$ denotes the number of joints of $i^{th}$ part. Then, we adjust the number of channels of part feature to $l$, where the $l$ denotes the number of spatio-temporal interaction layers. After that, we split channels into $l$ subgroups, generating spatial features $S$ with the shape of $T\times V\times 1$ for further spatio-temporal interaction.}

\subsubsection{Global\&Part-Level Spatio-Temporal Modeling} We achieve dual-level spatio-temporal modeling using the same pipeline for a more consistent and streamlined design. \new{At first, we utilize a simple 1D convolutional layer to adjust the first subgroup of $S$ to generate the initial temporal features $\mathcal{F}^{1}$. Then, multiple linear-transformer~\cite{Linformer1} layers are stacked to extract temporal features $\mathcal{F}^l \in \mathbb{R}^{C \times T}$ for each part, with formulation:
\begin{equation}
\mathcal{F}^{l} \leftarrow \mathrm{ReLU}[\phi(Q_t) \phi(K_t)^T V_t * W_t + \mathcal{F}^{l}], \text{w.r.t. }l>1.
\end{equation}}
Here, \(Q_t\), \(K_t\), and \(V_t\) are transformed from \(\mathcal{F}^{l} \) through linear layers \(W_{Qt},W_{Kt},W_{Vt} \in \mathbb{R}^{C \times C_t}\), where \(W_t \in \mathbb{R}^{C_t \times C}\) is a linear layer for channel adjustment, and \(\phi(\cdot)\) denotes the Sigmoid activation function.
The linear-transformer layer can capture the relationship across all frames, while keeping linear complexity. Then, we achieve layer-wise spatio-temporal interaction by cross-attention operation. Given the temporal features $\mathcal{F}^l$ and the spatial sub-features $S^{l}$, the spatio-temporal features can be generated as follows:
\begin{equation}
	\mathcal{F}^l \longleftarrow \text{CrossAtt}(\mathcal{F}^l, S^{l}) * \mathcal{F}^l + \mathcal{F}^l,
\end{equation}
\begin{equation}
	\text{CrossAtt}(\mathcal{F}^l, S^l) = \text{Softmax} \left((W^f S^{l})(\mathcal{F}^l)^T\right).
\end{equation}
Compared to previous methods, the DPE extracts part-level features in parallel, where the part-independence are well-preserved. Simultaneously, the decoupled spatio-temporal modeling efficiently alleviates the over-smoothing issue.

\subsubsection{Part-Global Interaction} Despite the parallel extraction keeps the part independence, the global modeling is invisible to each branch. To enable the parts to capture the semantic prompt from the whole-body movement, we incorporate a part-global interaction module also implemented by the cross-attention operation. 
Given the whole-body feature $\mathcal{F}_g$, the interaction between the $i^{th}$ part and whole-body is formulated as:
\begin{equation}
	\mathcal{F}_i \longleftarrow \text{Softmax} \left((W^g \mathcal{F}_g) \mathcal{F}_i^T\right)\mathcal{F}_i + \mathcal{F}_i,
\end{equation}
where $W^g$ is a 1D convolution layer to adjust the dimension. This interaction helps each part capture the semantics of body motion for a more accurate understanding.

At last, we fuse the part features \new{$\mathcal{F}_I = \{\mathcal{F}_i | i\in[1,\dots,T]\}$} and whole-body feature $\mathcal{F}_g$ with formulation:
\begin{equation}
	\mathcal{F} = \text{Projection}(\text{Concat}(\mathcal{F}_I,\mathcal{F}_g)),
\end{equation}
where the Concat denotes the concatenation along the channel dimension and the Projection is achieved by a 1D convolutional layer.
The motion feature $\mathcal{F}$ combined the motion representations from multiple parts and whole-body motion, resulting a more comprehensive motion understanding.

\subsection{Language-assisted Distribution Alignment}
\begin{figure}
	\centering
	\includegraphics[width=0.5\textwidth]{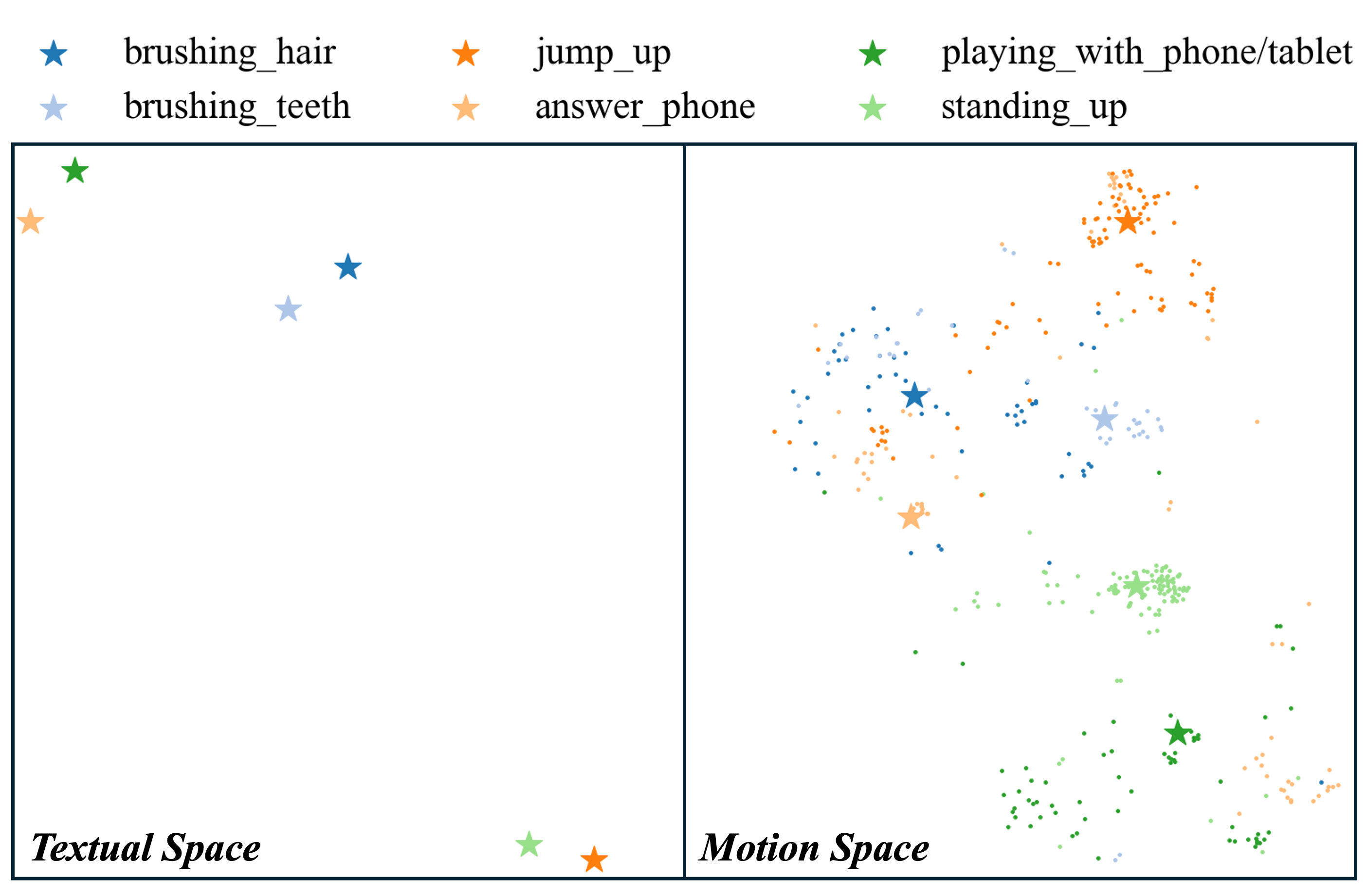}
	\caption{ \label{fig:text}The structure of representation space comparison between textual descriptions and the motions extracted from DPE (part \textit{hip} trained on Partial of PKU-MMD dataset). The \ding{72} in left figure denotes the textual representation extracted by CLIP(ViT/B-32)~\cite{CLIP}. The \ding{72} and $\bullet$ in right figure denotes the class-wise pooled representations and instance-wise segment representations, respectively. It can be visualized that the one-hot label optimized motion representation space exists 1) severe semantic correlations lost and 2) unclear inter-class decision boundary because of low intra-class compactness.}
\end{figure}

Although the DPE can extract more comprehensive motion representations from part and global views, we argue that the current general training pipeline leads to a sub-optimal motion representation space. For a better understanding, the comparison on structure of representation space between textual descriptions and the motions is illustrated in Fig.~\ref{fig:text}. Regarding textual space, we can get double observations. 1) The hip motion patterns are similar between homogeneous actions (e.g., \textit{Jump up} \& \textit{Standing up}) and 2) The degree of heterogeneity between different actions varies greatly (e.g., \textit{Brushing hair} \textit{vs.} \textit{Answer phone} and \textit{Brushing hair} \textit{vs.} \textit{Jump up}). These observations are highly aligned with human cognition, demonstrating the potential of the textual representations encoded by human language. However, in motion representation space, the homogeneity and the degree of heterogeneity is not well captured (e.g., \textit{Playing with phone} \& \textit{Answer phone}). Nevertheless, the segment representations perform low intra-class compactness (e.g., \textit{Answer phone}). It leads to unclear decision boundary between ambiguous actions (e.g., \textit{Brushing hair} vs. \textit{Brushing teeth}). To this end, the LDA aims to 1) supply cluster centers to improve the intra-class compactness and inter-class separation and 2) arrange cluster centers by semantic relationships. To achieve these goals, the proposed LDA acquires semantic-based cluster distribution with the help of large-scale language model at first, and aligns the skeleton representation distribution with the semantic-based distribution. Then we introduce the details of the implementation of LDA.
\subsubsection{Textual Distribution Aquisition} Given label definitions of $K$ actions, we prompt large-scale language model GPT-4~\cite{GPT-4} to generate the textual description of actions at body-level and part-level. For body-level description, we utilize the prompt ``\textit{Please describe the action [CLASS]}" where the [CLASS] denotes the name of actions. For part-level descriptions, we utilize the prompt ``\textit{Please describe the movement of [PART] when doing [CLASS]}" where The [PART] denotes the part definition given by the mapping $Q$. After that, we utilize a text encoder pretrained on large-scale text dataset, e.g., BERT~\cite{BERT} and CLIP~\cite{CLIP}, to extract textual features. Given the text descriptions, we generate text embeddings \new{$T=[T_g, T_1, \dots, T_I]$ where the $\{T_i | i\in [1,I]\}$} denotes the textual embedding of the $i^{th}$ part and the $T_g$ denotes those of whole body. 

\subsubsection{Skeleton-Text Alignment Loss} Taking the text embeddings as cluster centers, we motivate the body-level and part-level motion embeddings close to centers to arrange motion embedding space based on semantic correlation. Specifically, given body feature $\mathcal{F}_g$ and part features $\mathcal{F}_I$, we generate the segment embeddings by average pooling the action instances. Then we project the embeddings into a text-skeleton unified embedding space by a simple 1D convolutional layer. We compute the cosine similarities between embeddings and textual semantic centers and motivates the similarities between the same classes close to 1, and vice versa. Kullback–Leibler divergence (KL) loss is adopted between the cosine similarities matrix and the class identity matrix as $\mathcal{L}_{aln}$, formulated as follows:
\begin{equation}
	\hat{\mathcal{L}}_{aln}\! =\! \sum_i\! \left(\text{KL}(\cos(\mathcal{F}_i, T_i), \!\text{GT})\!\right)  + \text{KL}(\cos(\mathcal{F}_g, T_g), \!\text{GT}),
	\label{loss:aln}
\end{equation}
where the GT is the binary identity matrix derived from the ground-truth. 
Considering that Eq.~\ref{loss:aln} may lead to inconsistent alignment when the dataset is class-imbalanced, we incorporate a weight vector $w$ that depends on the number of instances within a batch. We generate class-wise alignment weight $w$ using:
\new{\begin{equation}
	w = \{\frac{\text{Count}(k)}{\text{Count}(instances)}, k\in [1, \dots K]\}.
\end{equation}}
Then we reformulate the Eq.~\ref{loss:aln} as follows:
\begin{equation}
	\mathcal{L}_{aln} = w * \hat{\mathcal{L}}_{aln}
	\label{loss:aln2}
\end{equation}

\subsection{Semantic Offset Adaptor} With Eq.~\ref{loss:aln2}, the segment-level skeleton embedding distribution are close to the textual distribution. However, we argue that the static textual embedding is not enough because of the large semantic gap between the skeleton domain and textual domains. The generated textual distribution can not entirely represent the real distribution, and the hard alignment leads to an inflexible optimization of the embedding space. Therefore, we incorporate a simple adapter to introduce semantic offset into static textual embeddings, making the optimization more flexible and more aligned with the real action distribution. We offer three choices for semantic offset learning:
\begin{itemize}
	\item \textbf{Prompt Learning.} Prompt learning methods argue that the traditional prompt ``\textit{This is a image of [CLASS]}" cannot fully realize the potential for alignment. Therefore, several studies have explored adding learnable tokens into prompts~\cite{tcsvt_prompt}. Derived from CoOp~\cite{zhou2022coop}, we add several (5 as default) learnable tokens into the generated text description for semantic offset learning. Taking \textit{Brushing hair} as an example. We set the description text as ``\textit{The hand moves up and down,} [TOKEN], [TOKEN], [TOKEN], [TOKEN], [TOKEN]". Then the prompt text is input into a textual encoder to generate textual features. During training, the learnable tokens are optimized in an end-to-end way.
	\item \textbf{Cross-domain Prompt.} To alleviate the domain gap between skeleton and text, we consider a cross-domain prompt as an option. Specifically, we apply two projection matrices  $W^K$ and $W^V$ on text embeddings and $W^Q$ on the skeleton embeddings to generate $K$,$V$, and $Q$, respectively. Then we compute cross-domain attention by $Att = \frac{QK}{\sqrt{d}}V$, where the $d$ denotes the dimension of $K$. We then achieve text features with formulation: $T = Att * T + T$. This interaction enables text features to retrieve motion patterns to help the alignment. 
	\item \textbf{Residual Adaption.} Fine-tuning a projection head has been proven to be an efficient way for downstream adaption~\cite{gao2024clip}. We build a simple MLP head which contains \{Linear, GeLU, Dropout, Linear\} to learn the semantic offset. We achieve final textual features by adding the original textual features and the learned semantic offset.
\end{itemize}

\subsection{Boundary Regression Branch}
Previous studies in TAS indicate that prediction smoothing based on action boundary detection can significantly improve the performance of models. Following~\cite{ishikawa2021alleviating}, we introduce a boundary prediction branch to classify if one frame is the action transition or not. Specifically, we apply a 1D convolution layer over the action representations as follows:
\begin{equation}
	P_b = \text{Sigmoid}(\text{Conv1D}(\mathcal{F})),
\end{equation}
where $P_b\in R^{1\times T}$ represents the probability that a frame is an action boundary. Then, multiple dilated convolutional stages are stacked to refine the boundary detection. The probability over a threshold (0.5 as default) will be treated as a boundary.
When inference, the predictions within two consecutive boundaries will be smoothed by majority voting. 

\subsection{Loss Function}
 Following previous TAS studies, we utilize the cross-entropy loss $\mathcal{L}_{ce}$ for framewise classification and  Gaussian similarity-weighted TMSE loss $\mathcal{L}_{gs-tmse}$ to suppress the hard transitions of action probabilities~\cite{ishikawa2021alleviating, DeST}. For boundary regression, a binary logistic regression loss $\mathcal{L}_{brb}$ is incorporated, formulated as
\begin{equation}
	\begin{split}
		\mathcal{L}_{brb} = \frac{1}{T} \sum_{t=1}^{T} \Bigl( & w_p B_b(t) \cdot \log Y_b(t)                   \\
		                                                      & + (1 - B_b(t)) \cdot \log (1 - P_b(t)) \Bigr),
	\end{split}
\end{equation}
where the $B_b(t)$ is the binary ground truth (1 on boundary) and $w_p$ is the weight of boundary samples. We combine the aforementioned losses as: 
\begin{equation}
    \mathcal{L}_{cls} = \mathcal{L}_{ce} + \alpha * \mathcal{L}_{gs-tmse} + \beta * \mathcal{L}_{brb}.
\end{equation}
Here the $\alpha$ and the $\beta$ are empirically set as 1.0 and 0.1 as default.
Overall, we integrate all losses formulated as:
\begin{equation}
	\mathcal{L} = \mathcal{L}_{cls} + \gamma * \mathcal{L}_{aln}
\end{equation}
\section{Experiment}
We experimentally verify the efficiency of the proposed LPL on three public skeleton-based action segmentation datasets, including Logistic Activity Recognition Challenge (LARa)~\cite{lara}, Traffic Control Gesture (TCG)~\cite{wiederer2020traffic} and PKU-MMD~\cite{liu2017pku}. 
\begin{itemize}
    \item \textbf{LARa}: LARa introduces typical warehouse activities, with fourteen individuals executing eight distinct actions. It utilizes an optical motion capture (MoCap) system that measures the three-axis positions and orientations of 19 limbs at 200 Hz. Each participant engaged in 30 two-minute sessions, resulting in 24,000 skeletal frames per session across 377 trials. Following previous studies, we down-sampled the sequence into 50 Hz~\cite{msgcn,DeST}.
    
    \item \textbf{TCG}: TCG aims to aid in the identification of traffic control gestures at European road intersections for autonomous driving. It employs a motion capture suit outfitted with IMU sensors across 17 body joints to document 15 gestures performed by five subjects, totaling 550 recordings. We adopt the cross-subject setting for evaluation.
    
    \item \textbf{PKU-MMD}: PKU-MMD is a densely annotated 3D human activity recognition dataset, featuring 66 participants who carried out 51 typical daily activities, including 41 individual and 10 interactive actions. Over 1000 long skeletal sequences were captured using three Kinect devices, each sequence lasting approximately 3--4 minutes and containing about 20 action instances. Evaluations are performed under cross-view and cross-subject settings.
\end{itemize}

\subsection{Evaluation Metrics}
Given $y=\{y_1,y_2,\dots,y_T\}$ where $y$ is the annotation for $T$ frames, the framewise predictions $\hat{y}=\{\hat{y_1},\hat{y_2},\dots,\hat{y_T}\}$ and $y$ is compared for evaluation.
\begin{itemize}
    \item \textbf{Accuracy} is defined as the proportion of frames correctly identified in a sequence, computed as \(Acc = {\sum_{t=1}^T(y_t = \hat{y_t})} / {T}\). This metric is sensitive to action boundaries shift and the predictions of long actions.
    
    \item \textbf{Edit Distance} assesses discrepancies between actual and predicted segments by calculating the minimum number of edits needed (insertions, deletions, substitutions) to convert one sequence into the other. It is normalized relative to the longer of the two segments.
    
    \item \textbf{F1 Score} measures the overlap between predicted and true action segments at the segment level using the intersection over union (IOU) with thresholds \( \tau \). It classifies segments as true positive or false positive based on whether the IOU exceeds \( \tau \). Then the score is computed with formula: \(F1 = 2 \times \frac{\text{prec} \times \text{recall}}{\text{prec} + \text{recall}}\). This measure is stable against minor temporal shifts but sensitive to over-segmentation errors. Thresholds \( \tau \) are set at 0.1, 0.25, and 0.5, where higher values represent more stringent criteria.
\end{itemize}

\subsection{Implementation Details}
All experiments are conducted on a single RTX 3090 GPU. The number of temporal layers and skeleton feature bank splits are set as 10 following~\cite{DeST}. The average pooling is utilized to aggregate joint features into parts. For the training settings, we use an Adam optimizer with a learning rate of 0.001 for LARa and TCG, and 0.0005 for PKU-MMD. We train the model for 100 epochs For LARa and TCG dataset, and the batch size is set at 6. The model is trained for 200 epochs on the PKU-MMD dataset with a batch size of 8. We adopt the SGD optimizer with learning rate 0.01 for the semantic offset adaptor. This learning rate is adjusted by cosine learning scheduler and reduced to 0.001 after 50 epochs. Note that our proposed LDA serves as a motion representation refinement module and will be discarded during inference. 
\new{The single validation is used for evaluation for PKU-MMD (X-sub), PKU-MMD (X-view), and LARa datasets. For TCG dataset, the 5-fold cross-validation is used for evaluation, and the averaged results are reported.} The code is available at https://github.com/hitcbw/LPL.
\begin{table*}[!ht]
    \scriptsize
    \centering
    \caption{Comparison with the state-of-art on LARa and TCG. $\dag$ indicates our implemented results.}
    \begin{tabular}{lllllllllllllll}
        \toprule
        &\multirow{2}{*}[-1.0ex]{Method}   & \multicolumn{7}{c}{LARa} && \multicolumn{5}{c}{\new{TCG}} \\
         \cmidrule{3-9}\cmidrule{11-15}
        &&FLOPS$\downarrow$&Params.$\downarrow$  & Acc$\uparrow$ & Edit$\uparrow$ & \multicolumn{3}{c}{F1@\{10,25,50\}$\uparrow$} && Acc$\uparrow$ & Edit$\uparrow$ & \multicolumn{3}{c}{F1@\{10,25,50\}$\uparrow$}  \\
        \midrule
        \multirow{6}{*}{\begin{sideways}\textbf{VTAS}\end{sideways}} 	 	 	 	  	 	 	 	 
        &MS-TCN\dag~\cite{MS-TCN}& 3.13G  & 0.52M & 65.8 & - & - & - & 39.6 && 85.5 & 67.7 & 70.0 & 67.3 & 59.1 \\
        &MS-TCN{++}\dag~\cite{MS-TCN++}& 5.23G & 0.88M &71.7 & 58.6  & 60.1 & 58.6 & 47.0& & 86.0 & 71.0 & 74.6 & 71.7 & 62.9 \\ 	 	 	 	 	 	 	 	 
        &ETSN\dag~\cite{ETSN}& 4.30G & 0.72M & 71.9 & 58.4 & 64.3 & 60.7 & 48.1 & & 83.5 & 62.2 & 69.0 & 66.0 & 58.2 \\
        &ASRF\dag~\cite{ishikawa2021alleviating}& 7.09G & 1.19M & 69.5 &60.2 &-& - &50.2& & 83.9 & 65.3 & 71.5 & 69.9 & 62.4\\
        &ASFormer\dag~\cite{asformer}& 7.72G & 1.26M & 72.2 & 62.2 & 66.1 & 61.9 & 49.2& &82.3& 69.1 & 71.3 & 67.7 & 57.6 \\  	 	 	 	 	 		 	 
        &LTContext\dag~\cite{LTContext}& 3.60G & 0.55M& 71.4 &62.5 &68.1& 64.3 &52.1& &86.9  &75.3  &79.1   & 75.9 & 67.4\\
        &C2F-TCN\dag~\cite{LTContext}&10.78G & 4.41M & 72.0 & 56.4 & 61.8 & 57.2 & 45.3& &85.3 & 63.0 & 67.4& 63.8&55.1  \\
        \midrule    	 	 	 	 	 	 	 	 
        &MS-GCN\dag~\cite{MS-GCN}& 31.73G & 0.84M & 65.6 & - & - & - & 43.6& & 86.6& 68.7& 70.9&69.7&59.4 \\
        \multirow{6}{*}{\begin{sideways}\textbf{STAS}\end{sideways}}&CTC~\cite{CTC}& - & - & 69.2 & - & 69.9 & 66.4 & 53.8& & -& -& -&-&- \\
        &ST-GCN + C2F-TCN~\cite{CTC}& 37.78G & 4.08M &72.9 & 58.6 & 63.8 & 59.4 & 47.5& & 87.7 & 65.4 & 69.0 & 65.2 & 57.8\\
        &STGA-Net~\cite{STGA-Net}& - & - &   70.4 & 65.4 & - & - & 53.3& & - & - & - & - & -\\ 	 	 	 	 
        &Chai et al.~\cite{MTSTGCN}& 108.87G & 2.58M & 73.7& 58.6 &63.8& 59.4& 47.6& &85.9& 68.9& 71.3& 68.3&60.2\\
        	 		 	  	 	 	 	 
        &DeST-{TCN}~\cite{DeST}& 6.71G& 0.93M& 72.6 & 63.7 & 69.7 & 66.7 & 55.8& & 87.1 & 73.7& 78.7 & 76.9&69.6\\
        &DeST-Transformer~\cite{DeST}& 7.71G & 1.10M & 75.1 &64.2 &70.3& 68.0 &57.7 &&88.1&	75.7&	80.0&	78.2&	71.2\\
        \midrule    	 	 	 	 
        &LPL(Ours)& 34.73G & 2.73M & \textbf{76.1} & \textbf{65.2} & \textbf{72.3} & \textbf{70.0} & \textbf{58.6}& & \textbf{88.8} & \textbf{77.4} & \textbf{81.9} & \textbf{80.0}& \textbf{73.8}\\
        \bottomrule
    \end{tabular}
    \label{sota1}
\end{table*}
\begin{table*}[t]
    \scriptsize
    \centering
    \caption{Comparison with the state-of-art on PKU-MMD (X-sub) and PKU-MMD (X-view). $\dag$ indicates our implemented results.}
    \begin{tabular}{llllllllllll}
        \toprule
        \multirow{2}{*}[-1.0ex]{Method}   & \multicolumn{5}{c}{PKU-MMD (X-sub)} & & \multicolumn{5}{c}{PKU-MMD (X-view)} \\
        \cmidrule{2-6}\cmidrule{8-12}
         & Acc & Edit & \multicolumn{3}{c}{F1@\{10,25,50\}} && Acc & Edit & \multicolumn{3}{c}{F1@\{10,25,50\}}  \\
        \midrule
        ST-GCN\dag~\cite{ST-GCN} & 66.2 & 34.6 & 37.3 & 33.4 & 24.7 && 66.5 & 35.4 & 38.6 & 33.9 & 24.9 \\
        MS-TCN\dag~\cite{MS-TCN} & 67.4 & 64.8 & 66.9 & 62.1 & 48.7 && 58.2 & 56.6 & 58.6 & 53.6 & 39.4 \\
        MS-TCN++\dag~\cite{MS-TCN++} & 66.0 & 66.7 & 69.6 & 65.1 & 51.5 && 58.4 & 56.7 & 58.7 & 53.2 & 38.7\\
        ETSN\dag~\cite{ETSN} & 68.4 & 67.1 & 70.4 & 65.5 & 52.0 && 60.7 & 57.6 & 62.4 & 57.9 & 44.3 \\
        ASRF\dag~\cite{ishikawa2021alleviating} & 67.7 & 67.1 & 72.1 & 68.3 & 56.8 && 60.4 & 59.3 & 62.5 & 58.0 & 46.1\\
        MS-GCN\dag~\cite{MS-GCN} & 70.1 & 66.2 & 69.4 & 66.0 & 52.6 && 65.3 & 58.1 & 61.3 & 56.7 & 44.1\\
        CTC~\cite{CTC} & 69.2 & - & 69.9 & 66.4 & 53.8 && - & - & - & - & - \\
        DeST-{TCN}~\cite{DeST} & 67.6 & 66.3 & 71.7 & 68.0 & 55.5 && 62.4 & 58.2 & 63.2 & 59.2 & 47.6\\
        DeST-Transformer~\cite{DeST} &70.3 &69.3 &74.5 &71.0 &58.7 &&67.3 &64.7 &69.3 &65.6 &52.0\\
        \midrule
        LPL(Ours) &\textbf{74.7} &\textbf{72.7} &\textbf{78.6} &\textbf{75.8} &\textbf{64.3}  &&  \textbf{70.0} & \textbf{67.7} & \textbf{73.3} & \textbf{70.0} & \textbf{58.5}\\
        \bottomrule
    \end{tabular}
    \label{sota2}
\end{table*}
\subsection{Comparison with State-of-the-art}
To comprehensively verify the efficiency, we compare the proposed LPL with existing state-of-the-art
methods, including temporal modeling methods~\cite{ETSN, MS-TCN, MS-TCN++, LTContext, asformer, ishikawa2021alleviating, c2f} and spatio-temporal modeling methods~\cite{MS-GCN,CTC,DeST,MTSTGCN,STGA-Net}. All the methods follow the best configuration from their reports. Nonetheless, we utilize FLOPS and Number of Parameters for model complexity evaluation on LARa Dataset. Because the FLOPS is related to the input dimension, we report the FLOPS by fixing the input as \new{$\mathbb{R}^{12\times6000\times19}$}.

The comparison results are demonstrated in Table~\ref{sota1} and \ref{sota2}. It can be observed that the proposed LPL outperforms all compared methods across all metrics. Surprisingly, we find that recent VTAS methods, such as LTContext~\cite{LTContext} and ASRF~\cite{ishikawa2021alleviating}, perform comparably to STAS methods based on C-ST modeling~\cite{MS-GCN,MTSTGCN}, even without the skeleton modeling backbone. At the same time, the DeST modeling methods, DeST-Transformer and DeST-TCN, perform a significant performance improvement than other STAS methods. It indicates that the DeST modeling can efficiently capture the motion clues while the C-ST modeling methods suffer from the over-smoothed features and don't fully realize the potential of graph networks. Compared to whole-body modeling methods, our proposed LPL achieves significant performance improvement. It demonstrates the necessity of part-level independent encoding and language-based prior knowledge. \new{To provide a comprehensive understanding of the performance advantages of our methods, the performance of LPL across multiple splits are compared to the state-of-the-art method DeST-Transformer. The paired t-tests on five metrics on TCG dataset indicate that our method significantly outperforms DeST-Transformer in terms of Accuracy $(p=0.0037)$, F1@0.25 $(p=0.0305)$, and F1@0.5 $(p=0.0068)$. However, no significant differences were observed in Edit distance $(p=0.0805)$ and F1@0.1 metrics $(p=0.1685)$. It suggests that the proposed strategy primarily enhances the discriminative capabilities of motion features, which improves the classification and temporal localization of ambiguous actions. However, it does not significantly contribute to understanding the temporal relationships of actions or to the recognition of simple movements, which are highly related to the Edit distance and F1@0.1, respectively.}

In terms of model efficiency, our proposed approach maintains comparable or less computational complexity with classical C-ST methods, such as MS-GCN (+9\% FLOPS), ST-GCN + C2F-TCN (-9\% FLOPS, -49\% Params.) and Chai et al (-213\% FLOPS), even if we extract part motion features independently. The efficiency benefits from the decoupled spatio-temporal modeling structure, which involves significantly fewer graph convolutional operations. However, the parallel part encoding results in significantly higher complexity compared to other DeST models (DeST-{TCN} and DeST-{Transformer}). We argue that these methods only employ whole-body joint modeling, which is not sufficient for accurate motion understanding. 
\subsection{Generalization on Previous STAS methods}
Our proposed LDA can be integrated into previous STAS methods without additional inference costs. To evaluate the generalization ability of LDA, we inject LDA into training phase of previous STAS methods~\cite{msgcn, MTSTGCN}. The performance comparisons are depicted in Table~\ref{tab:ldlgcn}. It can be observed that the skeleton-text distribution alignment can significantly improve the performance of current methods. It demonstrate the effectiveness and generalization of LDA. 
\begin{table}
	\centering
	\caption{\label{tab:ldlgcn}Performance when collaborate with LDA}
	\begin{tabular}{l|llllll}
		\toprule
		Method                        & Acc           & Edit          & \multicolumn{3}{c}{F1@\{0.1, 0.25, 0.5\}}                  \\
		\midrule
		MS-GCN~\cite{msgcn} & 67.4          & 54.2          & 60.4          & 55.7                                      & 43.7           \\
		MS-GCN~\cite{msgcn}+ LDA & \textbf{72.2}  & \textbf{59.7}  & \textbf{65.1}  & \textbf{60.7}            & \textbf{48.0 }  \\
  Gains&+4.8&+5.5&+4.7&+5.0&+4.3\\
		\midrule
		Chai et al.\cite{MTSTGCN} & 73.7          & 58.6          & 63.8         & 59.4   & 47.6\\
		Chai et al.\cite{MTSTGCN} + LDA & \textbf{74.5}  & \textbf{60.3}  & \textbf{67.7}  & \textbf{64.0}     & \textbf{51.9}  \\
  Gains &+0.8&+1.7&+3.9&+4.6&+4.3\\
		\bottomrule
	\end{tabular}
\end{table}

\subsection{Qualitative Analysis}
\begin{figure*}
	\centering
	\subfigure[LARa \# L03-S08-R17-A17-N01]{
		\includegraphics[height=0.18\textwidth]{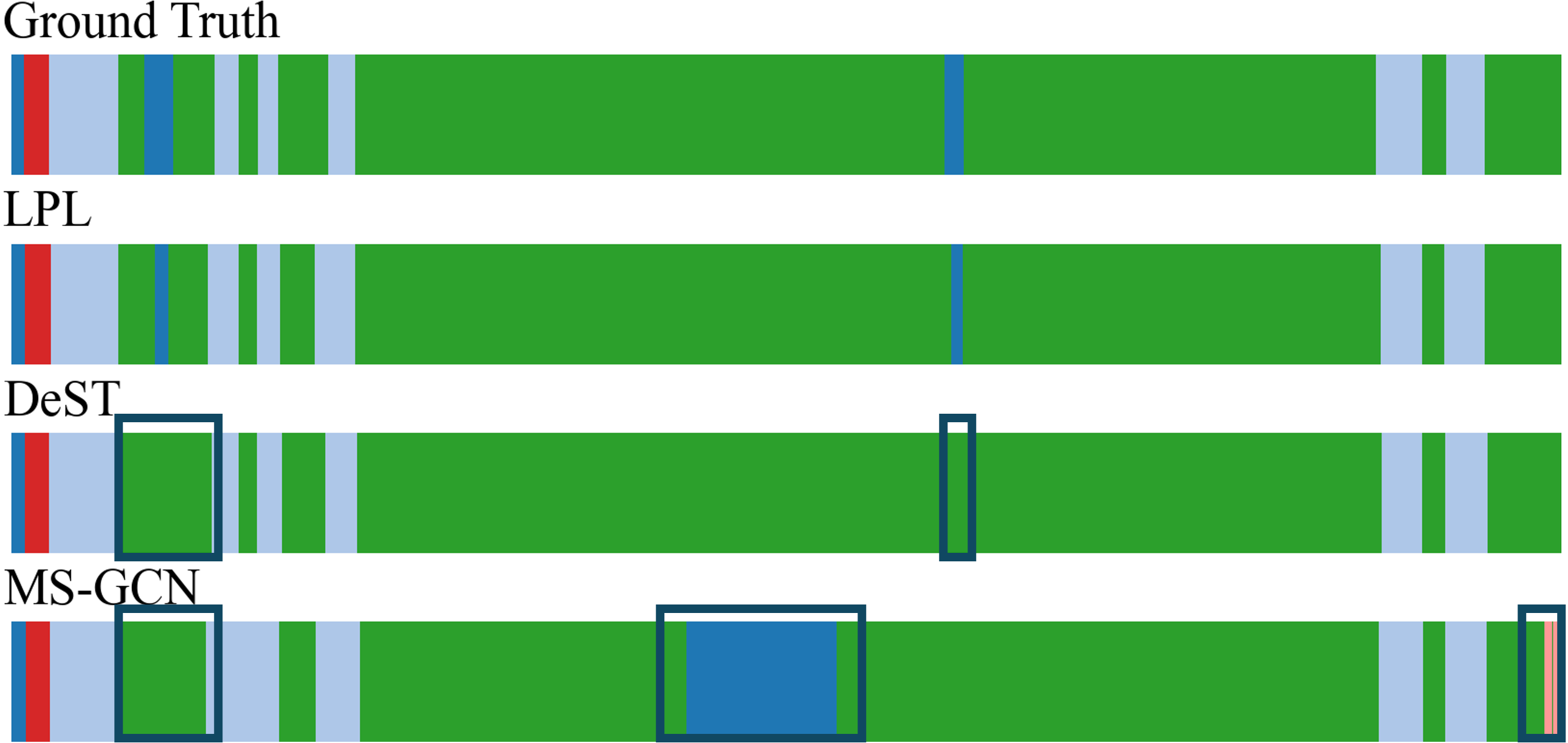}
	}
 \hspace{0.05\textwidth}
        \subfigure[LARa \# L03-S07-R18-A05-N01]{
		\includegraphics[height=0.18\textwidth]{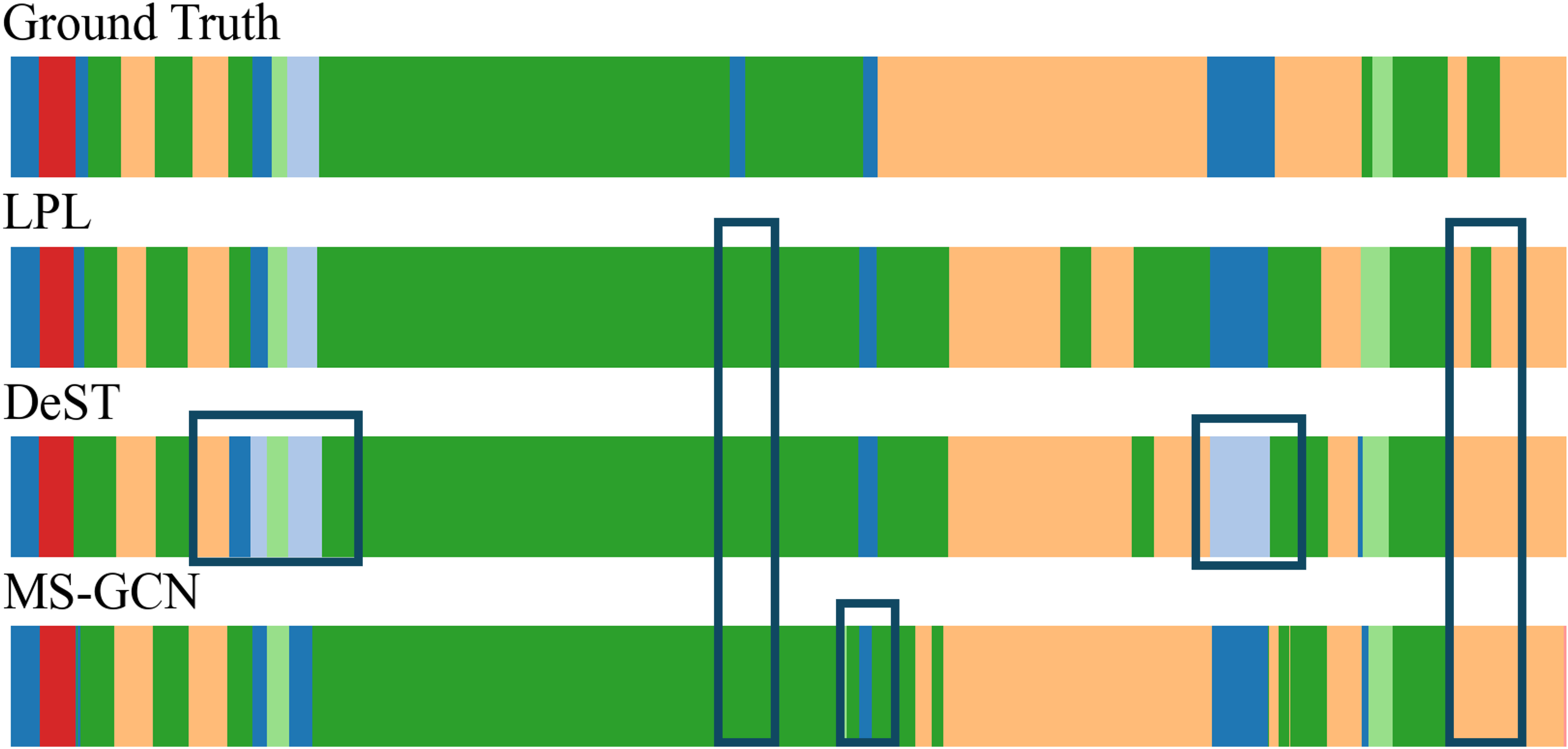}
	}
	\subfigure[PKU-MMD \# A18N07-L]{
		\includegraphics[height=0.18\textwidth]{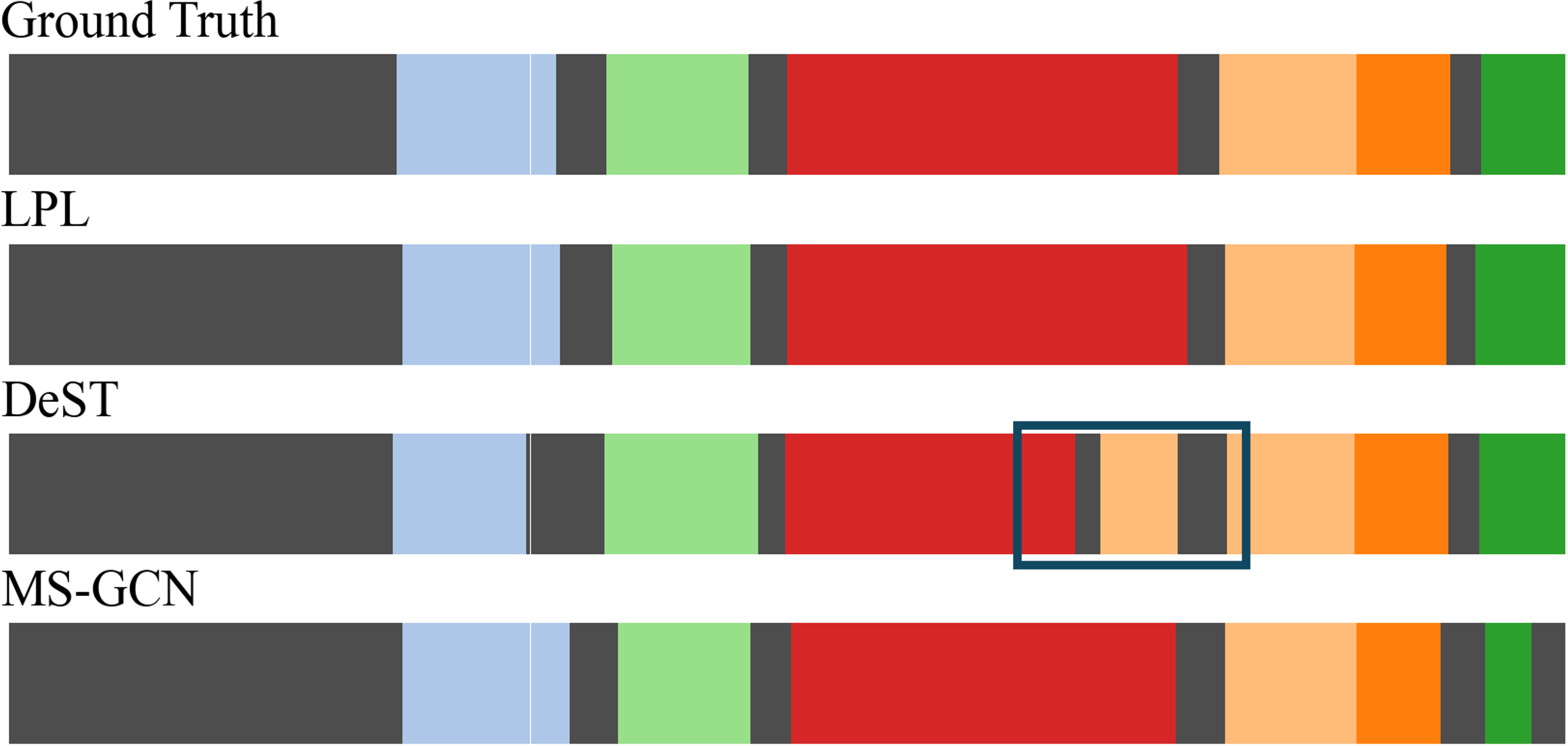}
	}
 \hspace{0.05\textwidth}
        \subfigure[PKU-MMD \# A16N03-L]{
		\includegraphics[height=0.18\textwidth]{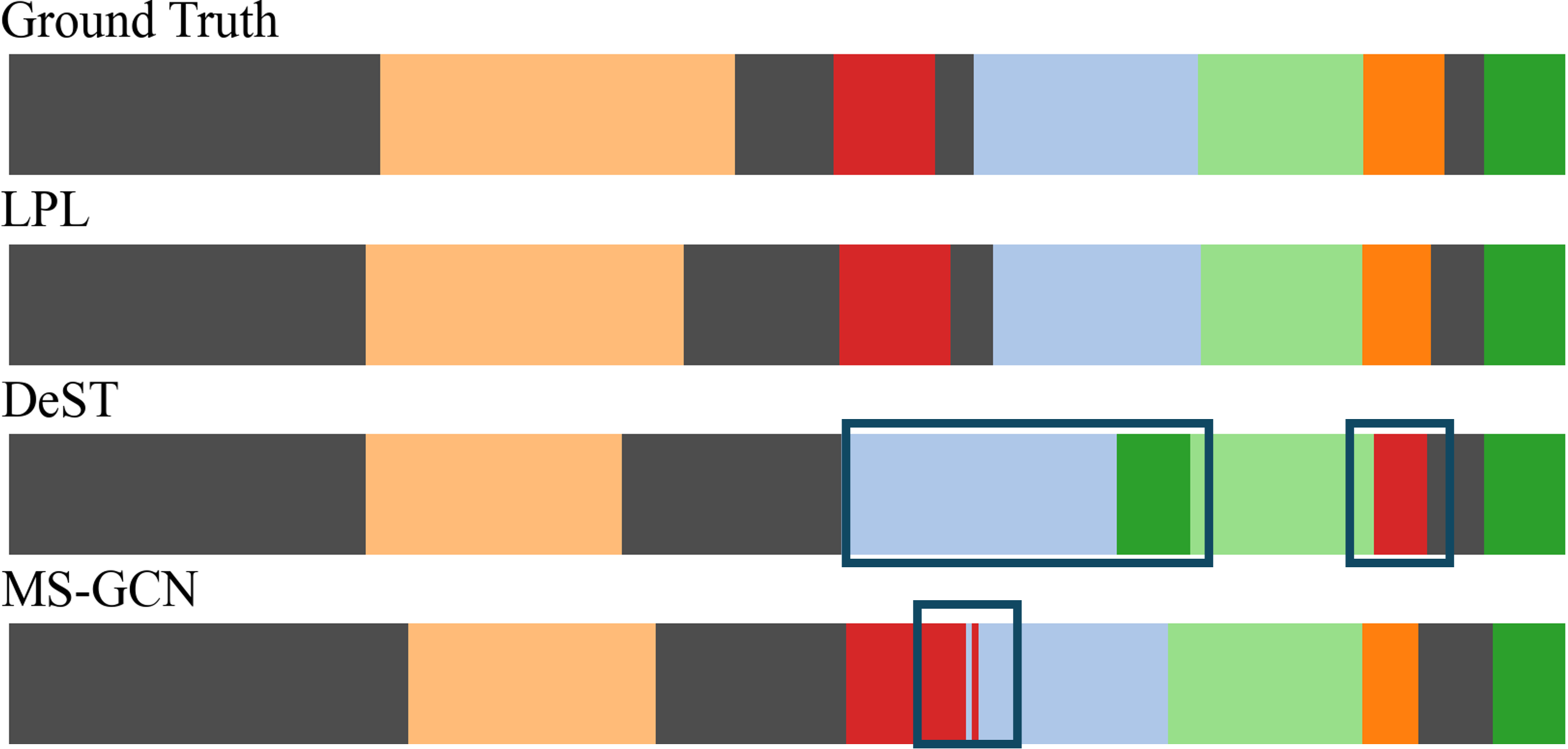}
	}
	\caption{\label{fig:qualiti} Predictions visualization from our LPL, DeST-Transformer, and MS-GCN. The first row shows the ground-truth labels, and the bottom three rows shows the corresponding predictions. The failure cases are enclosed by dashed boxes.}
\end{figure*}

We further conduct qualitative analysis on dataset PKU-MMD (x-subject) and LARa, as shown in Fig.~\ref{fig:qualiti}. Compared to the other two methods, the proposed LPL performs better segment retrieval ability, especially on identifying the \textit{Standing} instances colored blue, as visualized in in Fig.~\ref{fig:qualiti}(a) and (b). Also, LPL performs less wrong class identification when the action boundary is correctly identified, as illustrated in Fig.~\ref{fig:qualiti}(b) and (d). We hypothesize that the advantages come from stronger inter-class separation brought a better structure representation space. Compared to MS-GCN, the DeST-Transformer and the LPL performs less over-segmentation errors, highlighting the advantages of decoupled spatio-temporal modeling. 
\subsection{Ablation Study}
In this section, the effects of each component of LPL are experimentally discussed in detail. Model stability is also studied under different hyper-parameter choices.
To exclude the influence of other factors, all subsequent experiments use the same default settings except for the variables to be studied. \new{The first split of TCG dataset is selected for all the evaluation experiments.}
\subsubsection{Module Ablation}
\begin{figure*}
	\centering
	\subfigure[\textit{arm} in LPL]{
		\includegraphics[width=0.22\textwidth]{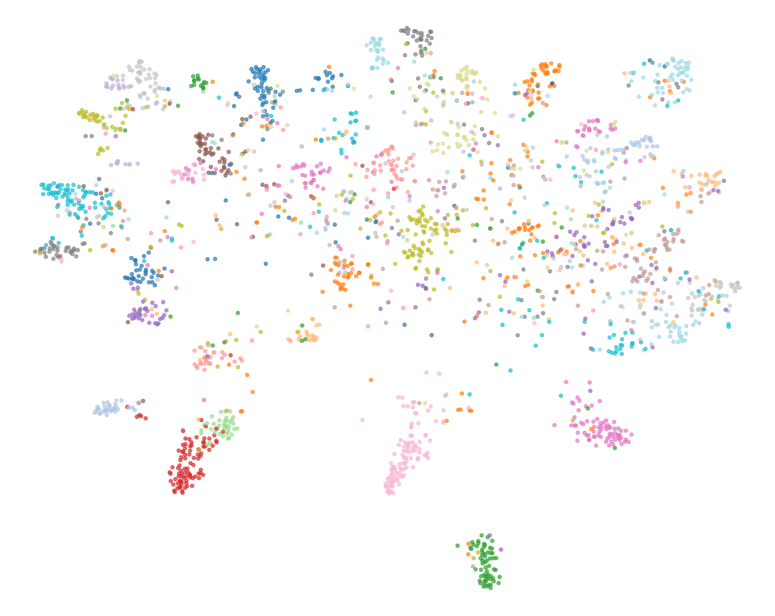}
	}
        \subfigure[\textit{leg} in LPL]{
		\includegraphics[width=0.22\textwidth]{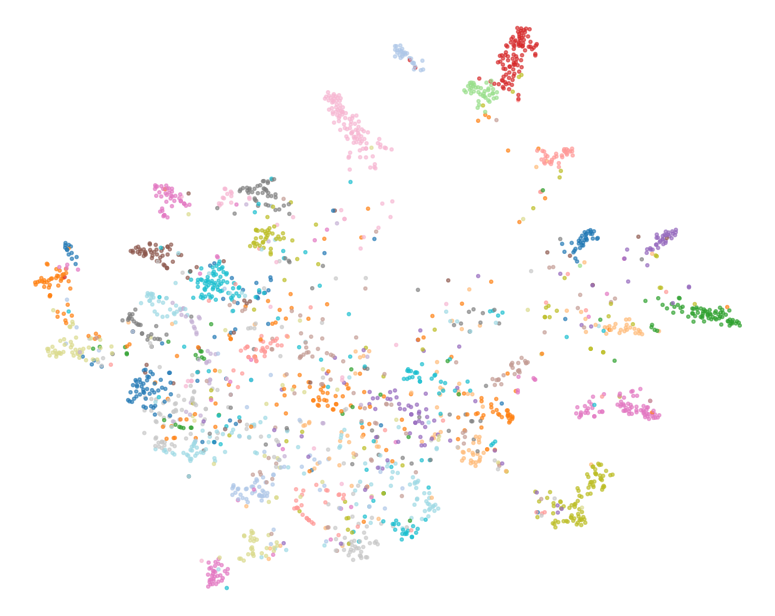}
	}
	\subfigure[\textit{fusion} in LPL]{
		\includegraphics[width=0.22\textwidth]{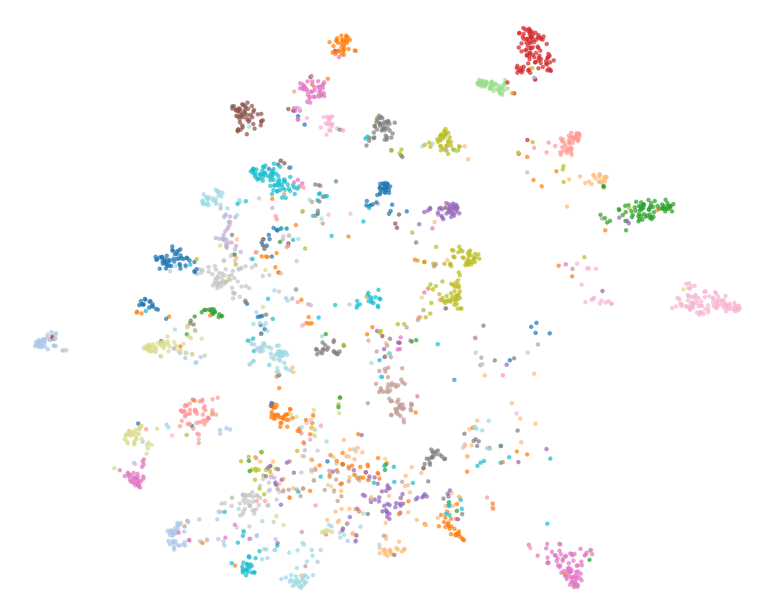}
	}
        \subfigure[\textit{fusion} in DeST-Transformer]{
		\includegraphics[width=0.22\textwidth]{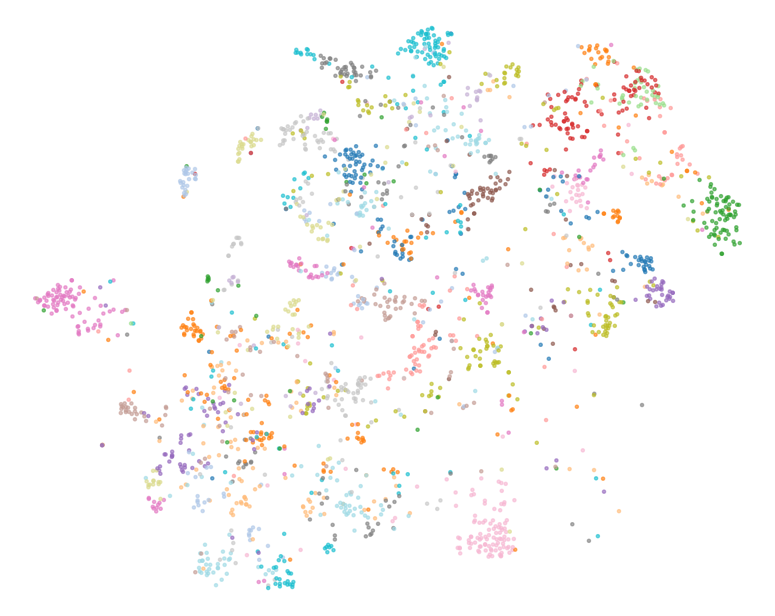}
	}
        \subfigure[\textit{arm} in DPE]{
		\includegraphics[width=0.22\textwidth]{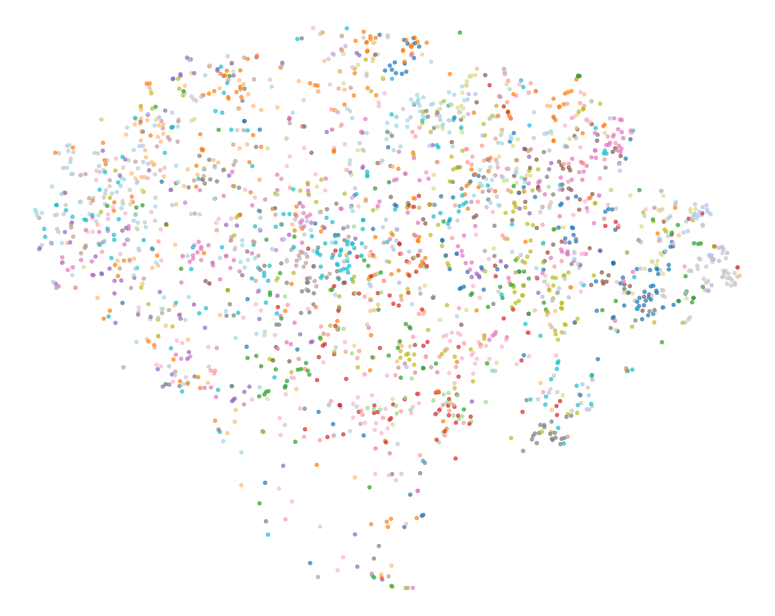}
	}
        \subfigure[\textit{leg} in DPE]{
		\includegraphics[width=0.22\textwidth]{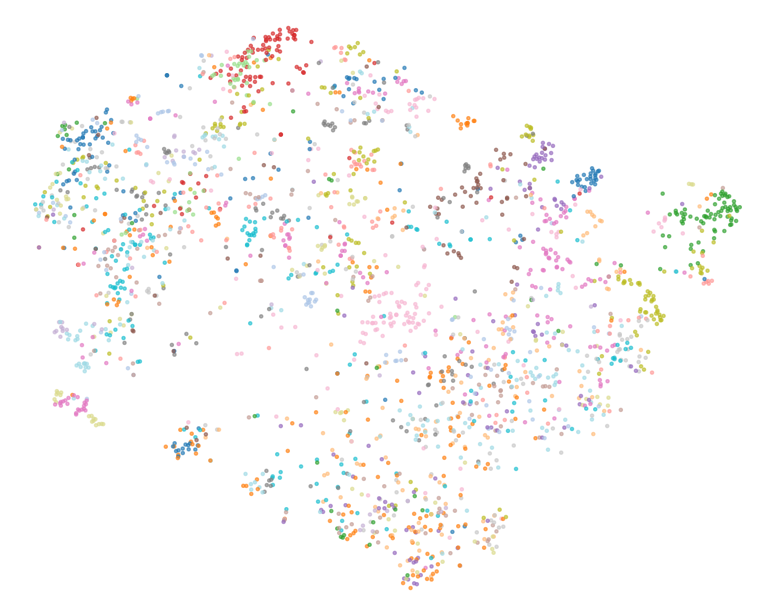}
	}
	\subfigure[\textit{fusion} in DPE]{
		\includegraphics[width=0.22\textwidth]{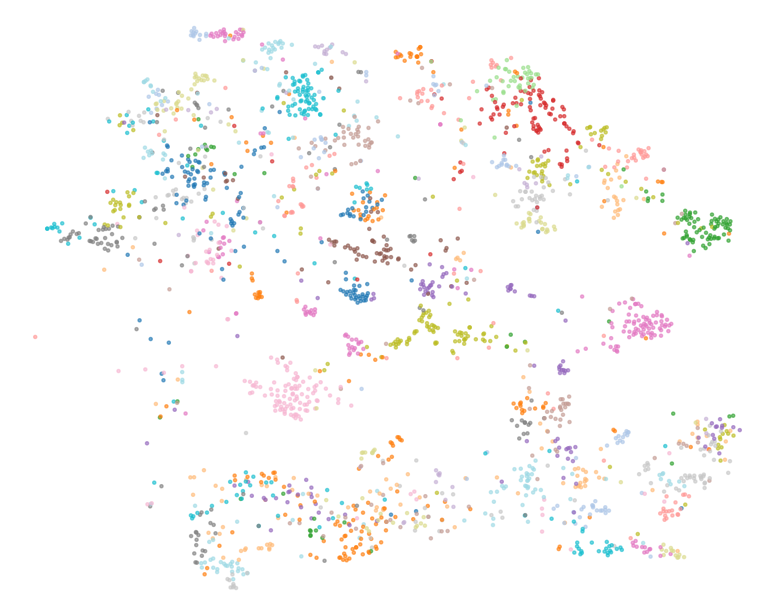}
	}
        \subfigure[\textit{fusion} in LPL w/o SOA]{
		\includegraphics[width=0.22\textwidth]{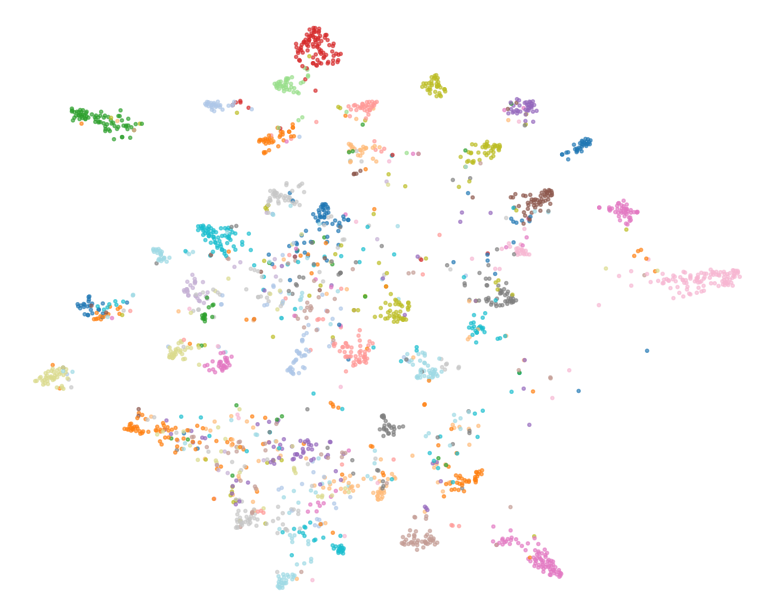}
	}
	\caption{\label{fig:repr} The visualized motion representation distribution for different parts. The \textit{fusion} denotes the fused representations from part-level and global-level features. It can be observed that the LPL owns clearer inter-class decision boundary than DeST-Transformer and DPE.}
 \end{figure*}

Our proposed LPL consists of the disentangled part encoder DPE and the skeleton-text alignment strategy LDA. The Table~\ref{abl:modules} presents the ablation study of these two modules. Compared to body-only modeling, the DPE fuse the motion perception from global-level and part-level, enabling better understanding of actions and slightly improving performance. Furthermore, the performance is significantly improved after the integration of the distribution alignment. To better understand the mechanism of LDA, we visualized the representation of action instances in Fig.~\ref{fig:repr}. Compared to DPE and DeST-Transformer, the part-level and global-level motion representations refined by LDA performs better spatial relationships. At first, the representations of same class behaves stronger intra-class compactness, and vice versa. Second, the representations of similar motions are close to each other, and vice versa. The intra-class compactness and inter-class separation helps the model learn more discriminative motion representations and achieve a clearer decision boundary. Moreover, LDA allows similarity among actions which share similar motion patterns at part-level, enable a comprehensive understanding of actions and avoiding focusing on noises when distinguish ambiguous actions. In summary, the LDA improves the intra-class compactness and inter-class separation from the view of part semantic, instead of easily push out representations only based on class identifications as previous contrastive learning methods~\cite{tcsvt_contra, tcsvt_contra2}. 
\begin{table}
	\centering
	\caption{\label{abl:modules}Ablation on Main Modules}
	\begin{tabular}{l|llllll}
		\toprule
		Method                        & Acc           & Edit          & \multicolumn{3}{c}{F1@\{0.1, 0.25, 0.5\}}                  \\
  \midrule
            \multicolumn{6}{l}{LARa}\\
		\midrule
		Baseline (DeST-Transformer) & 75.1          & 64.2          & 70.3          & 68.0                                      & 57.7           \\
  \midrule
		DPE & 74.5  & 64.7  & 71.2  & 68.2           & 57.3  \\
		DPE + $\mathcal{L}_{aln}$&     75.4      &   64.8     &71.5   & 68.8          & 57.8                \\
  DPE + SOA + $\mathcal{L}_{aln}$& 75.9          & 65.1         & 72.3 & \textbf{70.2}         & \textbf{58.2 }              \\
		DPE + SOA + weighted $\mathcal{L}_{aln}$& \textbf{76.1}          & \textbf{65.2}         & \textbf{72.3} & 70.0        & \textbf{58.6 }              \\
		\midrule
    \multicolumn{6}{l}{TCG}\\
		\midrule
		Baseline (DeST-Transformer) & 87.5&	76.2&	79.7&	76.8&	71.8           \\
  \midrule
		DPE & 87.5  & 73.3  & 79.7  & 78.6            & 73.2  \\
		DPE + $\mathcal{L}_{aln}$&  88.7          & 76.5        & 81.4          & 79.3                                      & 74.1           \\
  DPE + SOA + $\mathcal{L}_{aln}$& \textbf{88.9  }        & 77.8       & 81.7 & 79.0          & 74.4               \\
		DPE + SOA + weighted $\mathcal{L}_{aln}$& 88.7 & \textbf{78.5} & \textbf{82.0} & \textbf{79.6}& \textbf{75.0}          \\
		\bottomrule
	\end{tabular}

\end{table}
 
\subsubsection{The Number of Parts}
The part encoders within DPE extract the part-level motion representations in parallel. We then investigate the influence of the number of parts, and discuss whether all parts are necessary. We choose TCG dataset because its action definitions are gesture-based traffic signals, which are only related to the left and right arms, such as \textit{stop left static} and \textit{go right dynamic}. The performance comparison is illustrated in Table~\ref{abl:dpe}. It is evident that as the number of encoder branches increases, there is a general upward trend in model performance. It underscores the importance of disentangled part movement modeling and the language-assisted part motion representation refinement. We surprisingly find that the unimportant parts, i.e., the \textit{leg} and \textit{hip} also performs an important roles in motion understanding. In action description generation, the unrelated parts of these actions are described with the same texts. We suggest that even if these parts are not discriminative in action understanding, the LDA helps to pull these representations together and motivates the model to focus on more discriminative parts when classifying ambiguous actions.
\begin{table}
	\centering
	\caption{\label{abl:dpe}Ablation Study on Part Selection. The `b' denotes the body. The l\_ and r\_ denotes the left\_ and right\_.}
	\begin{tabular}{l|lllll}
		\toprule
		Method                        & Acc           & Edit          & \multicolumn{3}{c}{F1@\{0.1, 0.25, 0.5\}}                  \\
		\midrule
		b& 87.5&	76.2&	79.7&	76.8&	71.8      \\
		b + l\_arm &  88.8          & 76.3          &  81.0          & 77.8                                      & 72.4           \\
		b + l\_arm + r\_arm                 &  87.7          & 76.9          &  80.2          & 78.4                                      & 72.5          \\

		b + l\_arm + r\_arm + leg          & 88.6 & 77.6 & 81.2 &  78.3                             & 73.1 \\
		\midrule
		b + l\_arm + r\_arm + leg + hip                & \textbf{88.7} & \textbf{78.5} & \textbf{82.0} & \textbf{79.6}& \textbf{75.0}           \\
		\bottomrule
	\end{tabular}
\end{table}
\subsubsection{Implementation of Part Interactions and Semantic Adaption} For part-global interactions, we examined three implementation methods: addition, concatenation, and cross-attention. Table~\ref{abl:partglobal} shows that the cross-attention method yields the best performance, while addition performs comparably but with fewer parameters and less computational demand. We also investigated different implementations of Semantic Offset Adapters (SOA), as illustrated in Table~\ref{abl:soa}. We found that a simple residual adaptation effectively learns semantic offsets. Notably, the performance of the prompt tuning method varies across datasets: it achieves the best performance on the PKU-MMD dataset and the worst on the TCG dataset. We suppose that prompt tuning effectively harnesses the potential of the text encoder, but it may face challenges when the dataset scale is limited.
\begin{table}
	\centering
	\caption{\label{abl:partglobal}Different Implementations of Part-Global Interaction}
	\begin{tabular}{l|lllll}
		\toprule
  Implementation                      & Acc           & Edit          & \multicolumn{3}{c}{F1@\{0.1, 0.25, 0.5\}}                  \\
    \midrule
		None interaction & 75.6&	63.4&	70.8&	68.5&	57.7      \\
		add &   75.9     & 65.8 & 73.2 & 70.2 & 58.8  \\ 
		       concatenate          &  75.8          & 64.4          &  71.8          & 69.6          & 58.1         \\

		cross-attention          & 76.1 & 65.2 & 72.3 &  70.0                             & 58.6 \\
		\midrule
	\end{tabular}
\end{table}
\begin{table}
\caption{\label{abl:soa}Different Implementations of SOA}
    \centering
    \begin{tabular}{l|lllll}
    \toprule
    Implementation                      & Acc           & Edit          & \multicolumn{3}{c}{F1@\{0.1, 0.25, 0.5\}}                  \\
       \midrule
        \multicolumn{6}{l}{PKU-MMD (X-view Protocol)}\\
        \midrule
		Prompt Tuning & \textbf{70.4}&	67.5&	73.0	&\textbf{70.3}	&\textbf{59.0}      \\ 
		Cross-domain Prompt &  67.8          & 63.8          &  69.7          & 66.0                                      & 53.2           \\
		       Residual Adaption          &  70.0 & \textbf{67.7} & \textbf{73.3} & 70.0 & 58.5\\
                \midrule
        \multicolumn{6}{l}{TCG}\\
    \midrule
		Prompt Tuning & 89.0&	76.2&	80.6&	78.3	&73.1\\
		Cross-domain Prompt &  88.7          & 74.7          &  80.9          & 79.2                                      & 74.2           \\
		       Residual Adaption          &  \textbf{88.7}          & \textbf{78.5}         & \textbf{82.0} & \textbf{79.6 }         & \textbf{75.0}          \\
         \bottomrule
    \end{tabular}
    
    \label{tab:my_label}
\end{table}

\subsubsection{The choice of text encoder}
The text encoder plays a crucial role in determining the structure of the textual representation space within LDA. Existing text encoders are generally pretrained on large corpora, employing training approaches such as text-image alignment~\cite{CLIP} and token prediction~\cite{BERT}. We evaluated the effectiveness of using different text encoders for extracting textual representations and the comparison is illustrated in Table~\ref{tab:textencoder}. It can be observed that the CLIP text encoder achieved the best results.
\begin{table}
	\centering
	\caption{\label{tab:textencoder}Performance when collaborate with LDA}
	\begin{tabular}{l|llllll}
		\toprule
       
		Method                        & Acc           & Edit          & \multicolumn{3}{c}{F1@\{0.1, 0.25, 0.5\}}                  \\
  \midrule
   \multicolumn{6}{l}{LARa}\\
           \midrule
		
		CLIP~\cite{CLIP} & \textbf{76.1} & 65.2 & \textbf{72.3} & \textbf{70.0} & \textbf{58.6}           \\
		BERT~\cite{BERT} & 75.9  & 64.5  & 72.0   & 69.1 &58.5 \\ 
            T5~\cite{T5} & 75.9  & \textbf{65.5}  & 71.9  & 69.3     & 58.1\\
		\midrule
  \multicolumn{6}{l}{TCG}\\
  \midrule
  CLIP~\cite{CLIP} &\textbf{88.7} & \textbf{78.5} & \textbf{82.0} & \textbf{79.6}& \textbf{75.0}                                           \\
		BERT~\cite{BERT} & 87.8  & 76.0  & 80.7  & 79.2            & 74.0 \\
            T5~\cite{T5} & 88.6  & 75.8  & 80.4  & 78.7    & 73.1\\
            
            \bottomrule
	\end{tabular}
\end{table}
\subsubsection{Hyper-parameters}
We examine the impact of two critical hyper-parameters on model performance: the SOA learning rate \( l \) and the alignment loss weight $\gamma$. The weight $\gamma$ determines the alignment strength between motion representations and textual representations. A larger weight leads to stronger alignment and enhanced intra-class compactness, due to similar actions converging towards the same textual representation. However, due to significant variability in part movements among individuals performing the same action, excessive alignment might introduce some confusion into the model. The learning rate of SOA controls the intensity of semantic shifts during the current iteration; a learning rate that is too high may cause the model to overly focus on the samples of the current iteration, leading to excessive fluctuations in the learning process. Conversely, a learning rate that is too low can diminish the effectiveness of semantic shift adaptation. We present the performance comparision under different hyper-parameters in Table~\ref{tab:hyper}. It is observed that model performance is generally stable with respect to the alignment loss weight. Regarding the learning rate of SOA, excessive or insufficient learning rates result in reduced improvements or even negative effects.
\begin{table}
	\centering
	\caption{\label{tab:hyper}Hyper-parameters Ablation on LARa Dataset}
	\begin{tabular}{l|llllll}
		\toprule
       
		Value                       & Acc           & Edit          & \multicolumn{3}{c}{F1@\{0.1, 0.25, 0.5\}}                  \\
  \midrule
   \multicolumn{6}{l}{Learning rate of SOA}\\
           \midrule
		
		0.1 & 75.2 & 63.7 & 71.3 &68.8 & 57.0           \\
		0.01 & \textbf{76.1} & 65.2 & \textbf{72.3} & \textbf{70.0} & \textbf{58.6} \\ 
            0.001 & 75.6  & 64.6  & 71.8  & 68.2     & 57.7\\
		\midrule
  \multicolumn{6}{l}{Weight of $\mathcal{L}_{aln}$}\\
  \midrule
  0.1 &75.7 & 65.0 & \textbf{72.3} & 69.7& 58.1                                           \\
		0.5 & 75.8  & \textbf{65.3}  & 72.1  & \textbf{70.0}     & 58.2 \\
            1.0 & \textbf{76.1} & 65.2 & \textbf{72.3} & \textbf{70.0} & \textbf{58.6}\\
            \bottomrule
	\end{tabular}
\end{table}
\section{Conclusion}
In this article, we propose a language-assisted human part motion learning method for skeleton-based action segmentation, namely LPL. The key design of LPL lies in the part-aware encoder DPE and the skeleton-text distribution alignment strategy LDA. The DPE realizes part-level motion perception by parallel part encoders, while avoiding over-smoothness with decoupled spatio-temporal modeling. Simultaneously, the part-global interaction also enhances the comprehensive understanding of actions across multiple semantic granularities. Considering the information loss in one-hot encoded labels, LDA utilizes a large-scale language model to generate textual descriptions for each action. Leveraging the rich semantic information in the textual descriptions, LDA transfers the textual representation space distribution to the skeleton motion representation space thereby refining the structure of the motion representations. LDA not only improves intra-class separation but also organizes the representations from the perspective of human knowledge rather than simple action identification. The proposed LPL achieves state-of-the-art performance across various datasets. Moreover, the LDA can be incorporated into current STAS methods as an augmentation strategy for performance improvement.

\textbf{Limitations and Future Works.} Despite the proposed method achieving significant performance improvement, we believe that there is room for further refinement. Firstly, the proposed part encoder DPE realizes decoupled spatio-temporal modeling in a parallel way, which brings extensive computational overhead. Therefore, part-independent motion extraction in a computationally more efficient way is desired. Secondly, the global-part feature fusion in LPL is based on a simple concatenation, which we believe that the various part importance in classifying different actions are ignored. Efficient fusion of part features is worth further exploration.
\ifCLASSOPTIONcaptionsoff
	\newpage
\fi
\bibliographystyle{IEEEtran}
\bibliography{ref}

% \begin{IEEEbiography}{Michael Shell}
% 	Biography text here.
% \end{IEEEbiography}

% % if you will not have a photo at all:
% \begin{IEEEbiographynophoto}{John Doe}
% 	Biography text here.
% \end{IEEEbiographynophoto}

% \begin{IEEEbiographynophoto}{Jane Doe}
% 	Biography text here.
% \end{IEEEbiographynophoto}

\end{document}